\documentclass[10pt,twocolumn,letterpaper]{article}

\usepackage[pagenumbers]{iccv}      %

\def\ourDataset{CoIN-Bench} %
\def\ourDatasetVQA{IDKVQA} %
\def\ourDatasetVQALong{I Don't Know Visual Question Answering} %
\def\ourMethodLongBF{\textbf{A}gent-user \textbf{I}nteraction with \textbf{U}ncer\textbf{T}ainty \textbf{A}wareness} 
\def\ourMethod{AIUTA} %

\def\ourTask{CoIN} %
\def\ourTaskLong{Collaborative Instance Object Navigation}
\def\ourTaskLongBF{\textbf{Co}llaborative \textbf{I}nstance object \textbf{N}avigation}

\def\suppmat{\textit{Supp. Mat.}}

\newcommand{\sceneDescription}[0]{S\xspace} %
\newcommand{\sceneDescriptionRefined}[0]{S_{refined}\xspace} %
\newcommand{\sceneDescriptionEnriched}[0]{S_{enriched}\xspace} %

\newcommand{\factsTargetObject}[0]{F\xspace} %
\newcommand{\listOfQuestion}[0]{Q\xspace}

\newcommand{\xmark}{\color{Maroon}{\ding{55}}}
\newcommand{\mycheckmark}{\color{ForestGreen}{\ding{52}}}
\newcommand{\atoa}{{a\rightarrow a}}
\newcommand{\atou}{{a\rightarrow u}}
\newcommand{\utoa}{{u\rightarrow a}}
\newcommand{\monolithic}{Monolithic\xspace}

\newcommand{\sr}{\texttt{SR}\xspace}
\newcommand{\spl}{\texttt{SPL}\xspace}

\definecolor{iccvblue}{rgb}{0.21,0.49,0.74}
\usepackage[pagebackref,breaklinks,colorlinks,allcolors=iccvblue]{hyperref}
\usepackage[dvipsnames]{xcolor}
\usepackage{colortbl} 
\usepackage{multirow}

\usepackage{pifont}
\usepackage{listings}
\usepackage{algorithm}      %
\usepackage{algpseudocode}  %
\usepackage[commentColor=gray]{algpseudocodex}
\usepackage{alphalph}
\usepackage{soul}
\usepackage{microtype}

\def\paperID{12687} %
\def\confName{ICCV}
\def\confYear{2025}

\title{Collaborative Instance  Object Navigation: \\Leveraging Uncertainty-Awareness to Minimize Human-Agent Dialogues}
\author{
    Francesco Taioli$^{1,2}$,\ \  
    Edoardo Zorzi$^2$,\ \ 
    Gianni Franchi$^{3}$, \ \ 
    Alberto Castellini$^{2}$,\ \ 
    Alessandro Farinelli$^{2}$, \\ 
    Marco Cristani$^{2}$,\ \ 
    Yiming Wang$^{4}$ \\
    \small{$^1$Polytechnic of Turin,\ \ 
    $^2$University of Verona,\ \ 
    $^3$U2IS, ENSTA Paris, Institut Polytechnique de Paris,\ \ 
    $^4$ Fondazione Bruno Kessler}\\
    {\tt\footnotesize {francesco.taioli@polito.it}, \{name.surname\}@univr.it, {gianni.franchi@ensta-paris.fr}, {ywang@fbk.eu}}\\
    {\tt\small \url{https://intelligolabs.github.io/CoIN}}
}

\begin{document}

\definecolor{codegreen}{rgb}{0,0.6,0}
\definecolor{codegray}{rgb}{0.5,0.5,0.5}
\definecolor{codepurple}{rgb}{0.58,0,0.82}
\definecolor{backcolour}{rgb}{0.95,0.95,0.92}

\lstdefinestyle{mystyle}{
    backgroundcolor=\color{backcolour},   
    commentstyle=\color{codegreen},
    keywordstyle=\color{magenta},
    numberstyle=\tiny\color{codegray},
    stringstyle=\color{codepurple},
    basicstyle=\ttfamily\footnotesize,
    breakatwhitespace=false,         
    breaklines=true,                 
    captionpos=b,                    
    keepspaces=true,                 
    numbers=left,                    
    numbersep=5pt,                  
    showspaces=false,                
    showstringspaces=false,
    showtabs=false,                  
    tabsize=1
}
\lstset{style=mystyle}

\newcolumntype{s}{>{\columncolor[gray]{.85}[.5\tabcolsep]}c}
\twocolumn[{%
\renewcommand\twocolumn[1][]{#1}%
\maketitle
\begin{center}
\vspace{-0.65cm}
    \centering
    \captionsetup{type=figure}
    \includegraphics[width=1\linewidth]{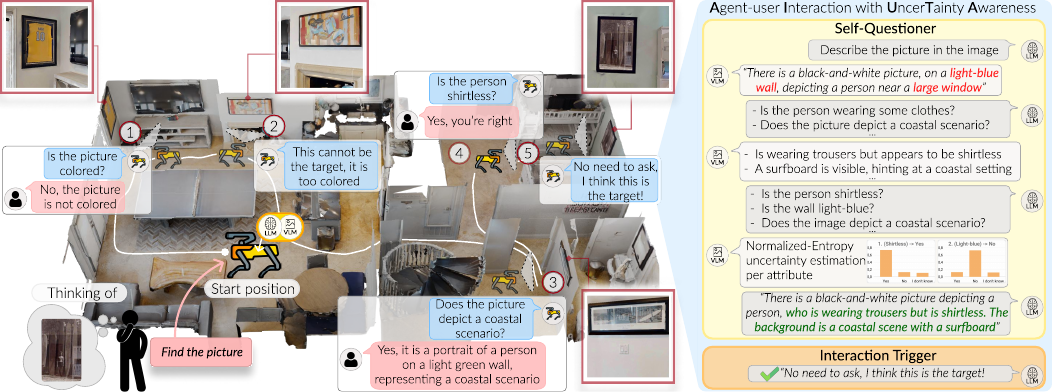}
    \caption{
    Sketched episode of the proposed \ourTask\ task.
    The human user (bottom left) provides a request (\emph{``Find the picture"}) in \textit{natural language}. 
    The agent has to locate the object in a \textit{completely unknown} environment \textit{without any target image as input}, interacting with the user only when needed via \textit{template-free}, \textit{open-ended} \textit{natural-language dialogue}.
    Our method, \textbf{A}gent-user \textbf{I}nteraction with \textbf{U}ncer\textbf{T}ainty \textbf{A}wareness~(\textbf{\ourMethod}), addresses this challenging task,
    minimizing user interactions by equipping the agent with two modules:~a \textbf{Self-Questioner} and an \textbf{Interaction Trigger},
    whose output is shown in the blue boxes along the agent's path (\ding{172} to \ding{176}), and whose inner working is shown on the right.
    The \textbf{Self-Questioner} leverages a LLM and VLM in a self-dialogue to initially describe the agent’s observation, and then extract additional relevant details, with a novel entropy-based technique to reduce \textbf{\textcolor{red}{hallucinations and inaccuracies}}, producing a refined~\textbf{\textcolor{ForestGreen}{detection description}}.
    The \textbf{Interaction Trigger} uses this refined description to decide whether to pose a question to the user (\ding{172},\ding{174},\ding{175}), continue the navigation (\ding{173}) or halt the exploration (\ding{176}).}

    \label{fig:teaser}
\end{center}%
}]
\begin{abstract}
Language-driven instance object navigation assumes that human users initiate the task by providing a detailed description of the target instance to the embodied agent.
While this description is crucial for distinguishing the target from visually similar instances in a scene, providing it prior to navigation can be demanding for human.
To bridge this gap, we introduce Collaborative Instance object Navigation (CoIN), a new task setting where the agent actively resolve uncertainties about the target instance during navigation in natural, template-free, open-ended dialogues with human.
We propose a novel training-free method, Agent-user Interaction with UncerTainty Awareness (AIUTA), which operates independently from the navigation policy, and focuses on the human-agent interaction reasoning with Vision-Language Models (VLMs) and Large Language Models (LLMs). 
First, upon object detection, a Self-Questioner model initiates a self-dialogue within the agent to obtain a complete and accurate observation description with a novel uncertainty estimation technique.
Then, an Interaction Trigger module determines whether to ask a question to the human, continue or halt navigation, minimizing user input.
For evaluation, we introduce CoIN-Bench, with a curated dataset designed for challenging multi-instance scenarios. 
CoIN-Bench supports both online evaluation with humans and reproducible experiments with simulated user-agent interactions. On CoIN-Bench, we show that AIUTA serves as a competitive baseline, while existing language-driven instance navigation methods struggle in complex multi-instance scenes.
Code and benchmark will be available upon acceptance.
\end{abstract}
    
\section{Introduction}
\label{sec:intro}
Recent advances in Large Language Models (LLMs) and Vision-Language Models (VLMs) have significantly reinvigorated research on \textit{language-driven} navigation tasks~\cite{anderson2018vision,taioli2024mind,an2024etpnav,li2021ion, goat_bench}, where human engages with embodied agents via natural language only, the most intuitive human-agent interaction among other forms (\eg, visual reference~\cite{imageNav}).
In this paper, we focus on the \textit{language-driven} Instance Object Navigation (InstanceObjectNav) task~\cite{li2021ion, goat_bench}, a practical task where the agent aims to locate a \textit{specific} instance within an unknown 3D scene, based on a detailed instance description (differently from ObjectNav~\cite{object_goal_nav} where \textit{any} object of a category can be located). 
The instance description typically contains nuanced details about the intrinsic (\eg, color, material) and extrinsic (\eg, context, spatial relations) attributes of the searched object instance, which are essential for \textit{uniquely identifying} the target amid visual ambiguity. 
However, the standard language-driven InstanceObjectNav task assumes that the detailed instance description is provided upfront, before navigation begins.
This assumption can be demanding and impractical in real world, as users may not be able or willing to supply all details in advance.

We introduce the \ourTaskLongBF{} (\textbf{\ourTask{}}) task, which engages a human user via natural-language dialogues to resolve instance visual ambiguity during navigation.
\ourTask{} enables human users to initiate the InstanceNav task \textit{without} providing extensive instance description. 
For instance, the user can just specify the instance category, \eg, ``\textit{Find the picture}'', a challenging minimal-guidance scenario.
Notably, CoIN introduces, for the first time, \textit{template-free}, \textit{open-ended} human-agent dialogues, a significant departure from the templated question-answer pairs used in prior work~\cite{alex_arena_nips}. 
Instead, our agent engages in dialogue solely based on the understanding gained during navigation.
Within \ourTask{}, two key research questions arise: \textit{1)} \textit{When} and \textit{2)} \textit{How} should agent-user interaction occur? To address the ``\textit{When}'', the agent must develop an internal model of its perceived environment to determine the optimal moments for seeking assistance from the user, resolving ambiguities effectively. To address the ``\textit{How}'', the agent must formulate \textit{the most informative questions} to maximize its chances of locating the target. 

We introduce a novel \textit{zero-shot} approach called \textbf{A}gent-user \textbf{I}nteraction with \textbf{U}ncer\textbf{T}ainty \textbf{A}wareness (\textbf{\ourMethod{}}). AIUTA equips the agent with two onboard modules, the \emph{Self-Questioner} and the \emph{Interaction Trigger}, leveraging pre-trained VLMs and LLMs without additional training. 
The \emph{Self-Questioner} enables the agent to autonomously generate \textit{self-dialogues} to inquire additional \textit{target-relevant} details, and verify essential details with a \textit{novel technique for uncertainty estimation}.
As shown in Fig.~\ref{fig:teaser}, upon detection, the LLM first prompts the VLM to obtain an initial detection description which can be \textit{incomplete} and \textit{inaccurate}. 
To enrich with target-relevant details, the LLM further generates questions for the VLM, whose responses complement the initial description. However, since VLMs cannot guarantee accurate responses grounded in the visual counterpart~\cite{eyes_wide_shut, liu2024paying,how_easy_to_fool}, we further prompt the LLM to generate sanity-check questions about all relevant details (e.g., ``\textit{Is the wall light-blue?}''). We instruct the VLM's response to be either \texttt{Yes}, \texttt{No} or \texttt{I don't know}, proposing a novel \textit{Normalized-Entropy}-based technique to quantify the VLM uncertainty.
Finally, the \textit{Interaction Trigger} module leverages the LLM to predict an \textit{alignment score} between the refined detection description and the known target’s \textit{facts} acquired from previous agent-human dialogues, if any.
With the score, the module decides whether to continue navigation, terminate it, or ask human clarifying questions.

To evaluate \ourTask{}, we propose the first benchmark, \textbf{\ourDataset{}}, 
with a curated dataset that specifically focuses on the visual ambiguity challenge in \ourTask{}.
The dataset is created on top of the recent large-scale GOAT-Bench~\cite{goat_bench}, where we only consider episodes involving multiple instances in a scene, with high-quality visual observations on the target.
In total, \ourDataset{} consists of $1,649$ evaluation episodes, with on average five distractors (non-target instances of the same category) per episode. %
Our benchmark supports \textit{on-line} evaluation with humans, as well as reproducible evaluation via simulated user-agent interactions. We empirically show that the simulated user-agent interaction yields results that are in line with human evaluation.
With \ourDataset{}, \ourMethod, \textit{while being training-free}, outperforms state-of-the-art InstanceObjectNav methods that are trained on the dataset in the \textit{zero-shot} setting, in terms of success rate and path efficiency. 
Finally, to evaluate VLM uncertainty estimation in the context of \ourTask{}, we introduce the ``\ourDatasetVQALong{}" (\textit{\ourDatasetVQA{}}), with human annotations. On \ourDatasetVQA{}, our proposed \textit{Normalized-Entropy}-based technique outperforms recent competitors~\cite{energy_based_OOD}, being a more reliable uncertainty measure.

\noindent\textbf{Paper Contributions} are summarized as follows:
\begin{itemize}
    \item We introduce \textit{\ourTask{}}, a practical setting for InstanceObjectNav, enabling minimal human input via agent-human dialogues during navigation. 
    \item We propose \textit{\ourMethod{}}, a training-free method addressing \ourTask{}, using self-dialogues within the agent to reduce perception uncertainty and minimize agent-user interactions.
    \item We introduce a novel \textit{Normalized-Entropy-based technique} to quantify VLM perception uncertainty, along with a dedicated \ourDatasetVQA{} dataset, demonstrating its superior reliability compared to recent competitors.
    \item We introduce \textit{\ourDataset{}}, a new benchmark featuring the challenging multi-instance scenarios of \ourTask{}, supporting evaluation with both human and simulated user-agent interactions for reproducibility.

\end{itemize}

\begin{figure*}[t!]
    \centering
    \includegraphics[width=0.85\textwidth]{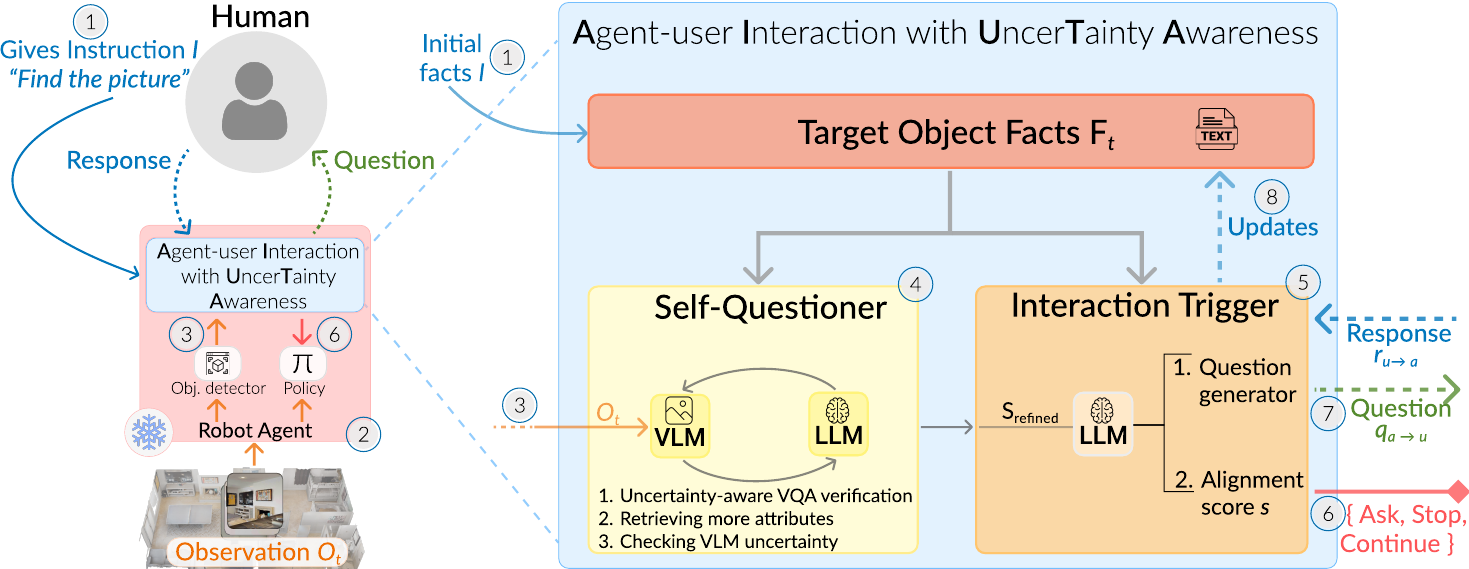}
    \caption{Graphical depiction of \textbf{AIUTA}: left shows its interaction cycle with the user, and right provides an exploded view of our method. \ding{172}~The agent receives an initial instruction $I$: ``Find a $c = <$object category$>$". \ding{173}~At each timestep $t$, a zero-shot policy $\pi$~\cite{yokoyama2023vlfm}, comprising a frozen object detection module~\cite{grounding_dino}, selects the optimal action $a_t$.
    \ding{174}~Upon detection, the agent performs the proposed AIUTA. Specifically, \ding{175} the agent first obtains an initial scene description of observation $O_t$ from a VLM. Then, a \textbf{\textit{Self-Questioner}} module leverages an LLM to automatically generate attribute-specific questions to the VLM, acquiring more information and refining the scene description with reduced attribute-level uncertainty, producing $S_{refined}$. \ding{176}~The \textbf{\textit{Interaction Trigger}} module then evaluates $S_{refined}$ against the ``\textit{facts}" related to the target, to determine whether to terminate the navigation (if the agent believes it has located the target object~\ding{177}), or to pose \textit{template-free, natural-language} questions to a human~\ding{178}, updating the ``\textit{facts}" based on the response~\ding{179}. }
    \label{fig:architecture}
\end{figure*}

\section{Related Works}
\label{sec:related_works}
\noindent\textbf{Instance Object Navigation.}
Instance Object Navigation (InstanceObjectNav) is an extension of Object-Goal navigation (ObjectNav)~\cite{anderson2018evaluation,batra2020objectnavevaluation}. 
Unlike ObjectNav, which seeks \textit{any} instance of a given category, InstanceObjectNav requires locating a specific instance defined by the user, making it a more practical and user-centered task. 
While the instance can be specified via an image (InstanceImageNav)~\cite{krantz2022instance}, we focus on the more intuitive setting where users describe the target instance only in natural language.
Recent policies can be divided into two categories: training-based~\cite{goat_bench, Ramrakhya_2023_PIRLNAV, zson, psl,Ehsani_2024_CVPR, yokoyama2024hm3d_ovon} and zero-shot policies~\cite{cows,esc,zhang2024trihelper, kuang-etal-2024-openfmnav, yokoyama2023vlfm,l3mvn}. 
Trained policies rely exclusively on reinforcement learning~\cite{goat_bench, psl, zson} or in conjunction with behavioral cloning~\cite{Ramrakhya_2023_PIRLNAV}.
Vision-language-aligned embeddings offer a promising alternative by enabling policies to incorporate detailed natural language descriptions as input.
For instance, GOAT-Bench~\cite{goat_bench} employ CLIP embeddings as the goal modality, while methods like~\cite{zson,psl} train on image-goal navigation~\cite{imageNav} and evaluate on the ObjectNav task. 
Among zero-shot policies, several methods extend the frontier-based exploration~\cite{frontier_based_expl}, by incorporating LLM reasoning~\cite{esc, l3mvn, zhang2024trihelper, kuang-etal-2024-openfmnav}, CLIP-based localization~\cite{cows} or vision-language maps for frontier selections~\cite{yokoyama2023vlfm}. The recent~\cite{barsellotti2024personalized} extends InstanceObjectNav to the context of personalization, providing the agent with multimodal instance references, \ie, a set of images and textual descriptions. Differently, we feature human-agent interactions during navigation, with no access to \textit{any} target image.

\noindent\textbf{Interactive Embodied AI.} 
Common approaches for human-agent interaction involve agents asking users for assistance, with responses typically consisting of shortest-path actions for reaching target objects~\cite{singh2022ask4help,just_ask} or simpler sub-goals expressed in natural language to guide navigation~\cite{help_anna,vision_based_nav, Liu_Paul_Chatterjee_Cherian_2024,avlen,rawal2024unmute}.
In~\cite{robot_ask_for_help_CORL_2023}, authors proposed a framework to measure the uncertainty of an LLM-based planner, enabling the agent to determine the next action or ask for help.
Both~\cite{alex_arena_nips, cvdn} include a dialog-guided task completion benchmark using human-annotated question-answer pairs collected via Amazon Mechanical Turk.
FindThis~\cite{majumdar2023findthis} requires locating a specific object instance through dialogue with the user.  
However, the agent only responds with images of candidate objects, lacking the ability to ask questions or engage in free-form natural language interactions, limiting its interactivity.
In~\cite{think_act_ask}, the Zero-Shot Interactive Personalized Object Navigation is proposed, where agents must navigate to personalized objects (\eg, ``Find Alice's computer"). 
However, personalized goals are manually annotated, and the user, simulated by an LLM, can only respond with this ground-truth data. 
Both ~\cite{majumdar2023findthis, think_act_ask} rely on a pre-built top-down semantic/occupancy map to locate the objects of interest; %
in contrast, our agent %
identifies target instance only through open-ended, template-free, natural language dialog with the user.

\noindent\textbf{Vision-Language Models Uncertainty.}
Hallucinations, biases, reasoning failures and the generation of unfaithful text by LLMs are well-known issues~\cite{llm_hallucination}.
Research by~\cite{orgad2024llms} shows that truthful information tends to concentrate on specific tokens, which can be leveraged to enhance error detection performance. However, these error detectors fail to generalize across datasets.
Similarly, recent studies highlight systematic limitations in the visual capabilities of large vision-language models~\cite{eyes_wide_shut, liu2024paying}, leading them to respond to unanswerable or misleading questions with hallucinated or inaccurate content~\cite{how_easy_to_fool}. To this end, PAI~\cite{liu2024paying} proposes adjusting and amplifying attention weights assigned to image tokens, encouraging the model to prioritize visual information. In~\cite{zhao2024first}, a linear probing on the logits distribution of the first tokens determines whether visual questions are answerable/unanswerable. A relevant work~\cite{zhu2024chatgpt} introduces a VLM-LLM dialogue for image captioning. Differently, \ourMethod{} leverages the self-dialogue for an embodied task, incorporating both observation and target facts for question generation, with a novel uncertainty estimation technique.

\section{\ourTaskLong}\label{sec:task}
\ourTaskLong~(\ourTask) introduces a novel setting for the InstanceObjectNav task, where an agent navigates in an unknown environment to locate a specific target instance in collaboration with a human user via \textit{template-free}, \textit{open-ended} and \textit{natural-language} interactions. The agent decides whether an interaction is needed to gather necessary target information from the user during the navigation.
The objective of \ourTask~is to successfully locate the target instance with \textit{minimal user input}, reducing the effort for the user in providing a detailed description.

Initially, the agent is positioned randomly in an \textit{unknown} 3D environment~\cite{hm3d_sem}. 
The navigation starts upon receiving a user request $I$ in natural language, %
which can be as minimal as by only specifying an open-set category $c$, \eg, ``\textit{Find the $<$category$>$}". 
The agent does not have access to any visual reference of the target instance.
We assume that the user is:~\textit{(i)}~aware of the full details about the target instance, and \textit{(ii)}~\emph{collaborative} to provide the true response when being asked by the agent.
At each time step $t$, the agent perceives a visual observation $O_t$ of the scene,  %
allowing it to %
guide a policy $\pi$ to pick an action $a_t %
\in A=\{$\texttt{Forward 0.25m, Turn Right 15°, Turn Left 15°, Stop, Ask}$\}$, 
where \texttt{Ask} is the novel action that comes with our \ourTask{} task.
When invoked, the agent asks the user a %
\textit{template-free open-ended} question $q_{\atou}$ in natural language to gather more information about the target. 
With the user response $r_{\utoa}$, the agent updates the set of \textit{facts} (set of attributes and characteristics) $\factsTargetObject_t$, representing information derived exclusively from the interaction.
Formally, the updated set of facts is represented as $\factsTargetObject_{t} = \factsTargetObject_{t -1} \cup r_{\utoa}$. 
The navigation terminates when certain criteria are met, \eg, the agent selects the \texttt{Stop} action or exceeds the maximum number of allowed actions. Notably, the agent can move anywhere in the continuous environment~\cite{Savva_2019_ICCV_habitat}.
\ourTask{} is particularly relevant in \textit{challenging} scenarios where many visually ambiguous instances co-exist.

\section{Proposed Method}

\label{sec:method}
Our proposed \ourMethodLongBF{}~(\textbf{\ourMethod}), a module that enriches the agent, is illustrated in Fig.~\ref{fig:architecture}.
Upon receiving an initial user request $I$ with minimal guidance that only specifies the category, \eg, \textit{``Find the picture"} (\ding{172} in Fig.~\ref{fig:architecture}), 
\ourMethod{}    updates the known facts regarding the target instance, \ie, $\factsTargetObject_{t= 0} = \{I\}$. Then, it activates a zero-shot navigation method, VLFM~\cite{yokoyama2023vlfm}, perceiving the scene observation $O_t$ and providing the navigation policy (\ding{173} in Fig.~\ref{fig:architecture}).
VLFM constructs an occupancy map to identify frontiers in the explored space, and a value map that quantifies the semantic relevance of these frontiers for target object localization using the pre-trained BLIP-2~\cite{blip_2} model.
Object detection is performed by Grounding-DINO, an open-set object detector~\cite{grounding_dino}. More details about VLFM~\cite{yokoyama2023vlfm} in the \suppmat{} (Sec.~\ref{supmat:vlfm}).

\ourMethod~is triggered upon the detection of an object belonging to the target class
(\ding{174} in Fig.~\ref{fig:architecture}), executing two key components sequentially. First, the \textit{Self-Questioner} (Sec.~\ref{sec:self-questioner}) leverages a Vision Language Model (VLM) and a Large Language Model (LLM) to obtain an accurate and detailed understanding of the observed object via self-questioning, enabling reliable verification of the detection against the target (\ding{175} in Fig.~\ref{fig:architecture}). 
Next, the \textit{Interaction Trigger} (Sec.~\ref{subsec:interaction_reasoner}), determines whether an agent-user interaction is necessary (in such case, triggering the action \texttt{Ask}), based on the observed object and known target facts $\factsTargetObject_{t}$, and whether the agent should halt (\ie, \texttt{Stop}) or proceed with the navigation (\ding{176} in Fig.~\ref{fig:architecture}). 
In the case of \texttt{Ask} (\ding{178} in Fig.~\ref{fig:architecture}), \ourMethod{} updates the target facts $\factsTargetObject_{t}$ with the response from the user (\ding{179} in Fig.~\ref{fig:architecture}). 
The agent terminates the navigation task once the target instance is deemed to be found.
The complete algorithm can be found in \suppmat{} (Sec.~\ref{supmat:algorithm}). In the following, \textit{Self-Questioner} and \textit{Interaction Trigger} are fully detailed.

\subsection{Self-Questioner}
\label{sec:self-questioner}
Upon detection, the Self-Questioner component aims to obtain a thorough and accurate description of the detected object. 
As suggested by previous studies~\cite{eyes_wide_shut, liu2024paying,how_easy_to_fool}, generative VLMs may produce descriptions that are not fully grounded on the visual content, leading to inaccuracy or hallucination. 
To mitigate this issue, we leverage an LLM to automatically generate attribute-specific questions for the VLM. In particular, we propose a novel technique for estimating uncertainty in VLM perception, enabling the refinement of detection descriptions. The technique has three steps:~\textit{(i)}~generating an initial detection description with detailed information relevant to target identification;~\textit{(ii)}~estimating VLM perception uncertainty to validate object detection; and~\textit{(iii)}~refining the detection description by filtering out uncertain attributes. Each step is detailed below.

\noindent\textbf{Generation of the initial detection description.}
The agent initially prompts the VLM for an initial description $\sceneDescription_{init}$ of the observation $O_t$ by providing the prompt $P_{init} =$ \textit{``Describe the \texttt{<}target object\texttt{>} in the provided image."} Formally,  
$\sceneDescription_{init}~=~\text{VLM}(O_t, P_{init})$. The description $\sceneDescription_{init}$ returned by the VLM could miss essential details for locating the specific instance, \eg, when looking for a picture, the content of the picture itself may not be specified in the description. 
To mitigate this issue, we prompt the LLM to create a list of questions $\listOfQuestion^{details}_\atoa  = \{ q_{j} \}$ %
 given $\sceneDescription_{init}$ and $\factsTargetObject_t$ (the symbol $\listOfQuestion_\atoa$ is used to represent the self-dialogue performed by the \textit{agent}). Formally,
 $\listOfQuestion^{details}_\atoa~=~\text{LLM}( P_{details},\sceneDescription_{init}, \factsTargetObject_t)$, where $P_{details}$ is the prompt guiding the question generation to obtain more details (\suppmat{} Sec.~\ref{supmat:additional_information}).
The questions of $\listOfQuestion^{details}_\atoa$ %
are subsequently answered by the VLM. Specifically, it answers each question $q_j \in \listOfQuestion^{details}_\atoa$ %
with a response $r_j~=~\text{VLM}(O_t, q_j)$ %
given the observation $O_t$. 
Finally, we concatenate all responses to the initial detection $\sceneDescription_{init}$, obtaining an enriched detection description $\sceneDescriptionEnriched$.
\noindent{\textbf{Perception uncertainty estimation.}
VLMs can generate hallucinated or inaccurate content \cite{eyes_wide_shut, liu2024paying,how_easy_to_fool}, impacting the performance of \ourMethod{}. To address this, we propose a novel %
technique for estimating their perception uncertainty. Direct evaluation of this aspect is challenging and often requires architectural modifications. Instead, we employ a prompt-guided Shannon entropy-based method for effective assessment. Our goal is to measure the uncertainty \( u \in [0, 1] \) of the VLM in perceiving specific aspects of a given image through visual question answering:  %
the VLM answers to a specific question $q$ with a response $r$ and an associated uncertainty estimation $u$, \ie, $(r, u) = \text{VLM}(O_t, q)$. 
Following the notation from~\cite{liu2024paying}, we consider an auto-regressive VLM, where $\mathbf{X}_I$ is the image representation (\ie, image tokens), $\mathbf{X}_P$ is the prompt representation (\ie, prompt text tokens), and $\mathbf{X}_H$ is the history representation (token generated at previous time-steps). 
During inference, the VLM generates a conditional probability distribution $p$ over the vocabulary $\mathbf{y} \in \mathbb{R}^{w}$ at each time step, expressed as:
 \begin{equation}
\begin{aligned}
\mathbf{y} &\sim p_{\text{VLM}}( \mathbf{y} \mid \mathbf{X}_I, \mathbf{X}_V, \mathbf{X}_H ),  \\
&\propto \operatorname{softmax} \left( \mathrm{logit}_{\text{VLM}}(\mathbf{y} \mid \mathbf{X}_I, \mathbf{X}_V, \mathbf{X}_H) \right).
\end{aligned}
\end{equation}
Estimating the uncertainty of the VLM response is non-trivial as the VLM has an unbounded output space and its output probability distribution is over a (large) vocabulary of size~$w$. 
To address this issue, we leverage the standard instruction-tuning~\cite{llava_1} procedures for VLMs, utilizing a predefined set of templated answers to restrict the vocabulary size to a fixed, small~$w$.
In particular, during inference, we use the following prompt: \textit{``\texttt{<}Question\texttt{>}? You must answer with \texttt{Yes}, \texttt{No}, or \texttt{?=I don't know}."}
In this way, we:~\textit{(i)} bound the auto-regressive nature to be essentially a one-step prediction, thus avoiding length-normalization; \textit{(ii)} bound the vocabulary size, \ie, $w=3$.
We then compute the Shannon entropy~\cite{shannon_entropy} $H$ of a probability distribution $p$ over vocabulary size $w$:
\begin{equation}
H(p_{\text{VLM}}) = - \sum_{i=1}^{w} p(y_i) \log p(y_i).
\end{equation}
The VLM uncertainty \( u \) is then obtained by normalizing the entropy \( H \) within the range \([0,1]\) as \( u = \frac{H}{H_{\text{max}}} \), where $H_{max} = \log(w)$ is the maximum entropy (\ie, maximum uncertainty) over a vocabulary of size $w$. 

Given a threshold $\tau$, we can indicate if the answer is \texttt{Certain} or \texttt{Uncertain}, namely:
\begin{equation}
\label{eq:abstein}
C(u,\tau) = 
    \begin{cases}
    \texttt{Certain}, & u \leq \tau \\
    \texttt{Uncertain}, & u > \tau
    \end{cases}
\end{equation}
To reduce false positives, we use the prompt $P_{check} = $ \textit{``Does the image contain a \texttt{<}target object\texttt{>}? Answer with Yes, No or ?=I don't know."} (see \suppmat{} Sec.~\ref{supmat:check_detection}).  
This allows us to confirm the presence of the object, which we formally express as $(r_{check}, u_{check}) = \text{VLM}(O_t, P_{check})$. Following Eq.~\ref{eq:abstein}, we continue the \ourMethod{} pipeline if response $r_{check} = $ \textit{``Yes"} and uncertainty $u_{check} =$  \texttt{Certain}; otherwise, we continue exploring.

To remove uncertain attributes, we prompt the LLM to extract a set of attributes and values $K_t = \left\{ (k_j,v_j)\right\}$ from the detection description $\sceneDescriptionEnriched$, where each attribute $k_j$ is associated to a value $v_j$, \eg, 
(``frame'', ``black''); (``content'', ``RGB image of a family''), etc.
For each attribute $k_j$, we then prompt the LLM to generate a list of $J$ questions, %
$\listOfQuestion^{attribute}_\atoa = \{ q_{j} \}_{j=1}^J$ to be answered by the agent itself.
Formally, we extract attributes list and self-questions in one prompt, 
$\listOfQuestion^{attribute}_\atoa~=~\text{LLM}(P_{self questions}, F, S_{enriched})$,
where $P_{self question}$ is the prompt for the LLM (\suppmat{} Sec.~\ref{supmat:self_questions)}). 
For each question $q_j$, we access both the response $r_j$ and the associated uncertainty $u_j$ by evaluating $(r_j, u_j)~=~\text{VLM}(O_t, q_j)$.
This process allows us to confirm or refine the attributes based on the VLM's responses, obtaining a final detailed description $\sceneDescriptionRefined$.

\noindent\textbf{Detection description refinement.}
To obtain the final detailed description $\sceneDescriptionRefined$, we let the LLM filter out uncertain attributes, given the enriched description $\sceneDescriptionEnriched$  and the set of questions, responses, and uncertainties $\{q_j, r_j, u_j\}$.
More formally, $S_{refined}=\text{LLM}(P_{refined}, \{q_j, r_j, u_j\}, S_{enriched})$, where $P_{refined}$ is the prompt for the LLM (see \suppmat{} Sec.~\ref{sup:mat:refined_image_desc}).

\subsection{Interaction Trigger}
\label{subsec:interaction_reasoner}
Using the accurate and detailed description $\sceneDescriptionRefined$ of the detected object, the Interaction Trigger prompts the LLM to decide whether to pose a question to the human user or continue the navigation. 
Specifically, we prompt the LLM to estimate a similarity score $s$ between scene description $\sceneDescriptionRefined$ and target object facts $\factsTargetObject_t$. 
We instruct the LLM to estimate the similarity score based on the alignment between the detection description and the known facts.
Formally, $s = \text{LLM} (P_{score}, \sceneDescriptionRefined, \factsTargetObject_t)$, where $P_{score}$ is prompt instructing the LLM to produce the similarity score (\suppmat{} Sec.~\ref{supmat:alignment_score}).
Based on the LLM-estimated similarity score, the agent takes corresponding action based on the following intuition:~\textit{(i)}~if $s \geq \tau_{stop}$, the navigation terminates as the agent deems the instance has been found;~\textit{(ii)}~if $s < \tau_{skip}$, the agent deems the detected object is significantly different from the known target facts, thus skipping the agent-user interaction to reduce the user efforts in providing input. The agent will continue with the environment exploration; 
and~\textit{(iii)}~if $ \tau_{skip} \leq s < \tau_{stop}$, the description and facts are somewhat aligned, suggesting that posing a question to the user can effectively reduce uncertainty.

When taking the action \texttt{Ask}, we further leverage the capability of LLM to compose an effective question to the user, $q_{\atou}$, aimed at maximizing information gain about the target instance, conditioned on the know target object facts $F$ and the refined observation description $\sceneDescriptionRefined$.
To minimize the number of LLM calls, we incorporate such question retrieval inside the $P_{score}$ prompt.
After receiving the corresponding response from the human, $r_{\utoa}$, we update the target object facts $\factsTargetObject_t$ with new information, maximizing the effectiveness of later agent-human interactions.

\section{\ourTask-Bench} \label{sec:benchmark}
To facilitate the evaluation of \ourTask{}, we introduce \textit{\ourDataset}, a curated dataset that features challenging multi-instance scenarios, supports both human evaluation and simulated agent-user interactions, and includes a new performance metric that accounts for agent-user interactions.

\noindent\textbf{Dataset Construction.}
Our dataset is built upon the large-scale GOAT-Bench~\cite{goat_bench}, which spans diverse scenarios from the HM3DSem~\cite{hm3d_sem} using the Habitat sim~\cite{Savva_2019_ICCV_habitat}. GOAT-Bench provides instance references in various formats, including category names and natural-language descriptions, making it a suitable source dataset.
GOAT-Bench consist of a large \texttt{Train} split for policy training, and three eval splits:~\texttt{Val Seen, Val Seen Synonyms} and \texttt{Val Unseen}. 
Specifically, \texttt{Val Seen} includes objects seen in \texttt{Train}, \texttt{Val Seen Synonyms} introduces synonymous object names, and \texttt{Val Unseen} contains only \textit{novel} objects absent from \texttt{Train}.
Since GOAT-Bench's \texttt{Train} split is dedicated to policy training, we design \ourDataset{} exclusively for evaluation. 

We select episodes from the evaluation splits of GOAT-Bench, \ie, \texttt{Val Seen, Val Seen Synonyms} and \texttt{Val Unseen} to ensure fair comparison with methods trained on GOAT-Bench. Since \ourTask{} focuses on scenarios with multiple instances of the same target category (\ie, \textit{distractors}), we apply a filtering procedure to discard episodes with fewer than $d_{min}=2$ distractors. 
After filtering the episodes, the simulator~\cite{Savva_2019_ICCV_habitat} sets random start positions to the agent, ensuring a geodesic distance of $[5m,20m]$ between the start and target locations to vary navigation difficulty.
Moreover, since visual observation are 3D renderings whose quality is dependent on the scene reconstruction, we manually filter out episodes to ensure high-quality visual observations, removing those where target instances have insufficient resolution, limited visual coverage, or are indistinguishable from distractors.
Additionally, we ensure episodes are navigable without crossing floors, following~\cite{habitatchallenge2023, goat_bench}.
\ourDataset{} dataset includes $831$ episodes in \texttt{Val Seen}, $459$ in \texttt{Val Unseen} and $359$ in \texttt{Val Seen Synonyms}, with a total of $1,649$ evaluation episodes, in line with the evaluation scale of well-known datasets~\cite{krantz2022instance, goat_bench,batra2020objectnavevaluation}. As shown in Tab.~\ref{tab:episode_counts}, \ourDataset{} features an average of $\sim5$ distractors per episode, and a mean path length $>7$, forming a highly challenging multi-instance evaluation set. More details and statistics are provided in \suppmat~(Sec.~\ref{supmat:ourdataset}).

\begin{table}[b]
    \centering
    \resizebox{1\columnwidth}{!}{%
    \begin{tabular}{lcccc}
        \toprule
        \multicolumn{1}{l}{Statistics} & \textit{Val Seen} & \textit{Val Seen Synonyms} & {Val Unseen}  \\
        \midrule
        Avg. (std) number of distractors& 4.58 (1.93)&  \multicolumn{1}{c|}{6.01 (1.96)}& 5.15 (1.51) \\
        Avg. (std) length (Geodesic)          &9.32 (3.43)&\multicolumn{1}{c|}{9.13 (3.14)}&9.86 (3.73)  \\ 
        Avg. (std) length (Euclidean)           &7.48 (2.88)& \multicolumn{1}{c|}{7.50 (2.75)}& 7.78 (3.39) \\
        \bottomrule
    \end{tabular}
    }
    \caption{Avg. (std) number of distractors and distance to the goal.}
    \label{tab:episode_counts}
\end{table}
\begin{table*}[ht!]
\centering
\resizebox{0.95\textwidth}{!}{
\begin{tabular}{rccc sccc sccc|c sccc}
    \toprule
    \multirow{2}{*}{Method \ \ \ \ \ \ \ \ \ \ \ \ \ \ \ } & \multicolumn{2}{c}{Model Condition} & \multicolumn{4}{c}{\textit{Val Seen}}  & \multicolumn{4}{c}{\textit{Val Seen Synonyms}}  & \multicolumn{1}{c}{} & \multicolumn{4}{c}{Val Unseen}  \\ 
    \cmidrule{2-3} \cmidrule{5-7} \cmidrule{9-11} \cmidrule{14-16} 
         & Input & Training-free && \scriptsize\textbf{SR}~$\uparrow$ & \scriptsize\textbf{SPL}~$\uparrow$ &\scriptsize\textbf{NQ}~$\downarrow$  && \scriptsize\textbf{SR}~$\uparrow$ & \scriptsize\textbf{SPL}~$\uparrow$ &\scriptsize\textbf{NQ}~$\downarrow$ &&& \scriptsize\textbf{SR}~$\uparrow$ & \scriptsize\textbf{SPL}~$\uparrow$ &\scriptsize\textbf{NQ}~$\downarrow$ \\ \midrule
          
         Monolithic$^\dagger$~\cite{goat_bench}~ (\textit{CVPR-24}) & d & \color{Maroon}{\ding{55}} &  &6.62$^\dagger$ & 3.11  &-    && 13.09$^\dagger$ & 6.45 & - &&& 0.22$^\dagger$ & 0.05 & -     \\

         PSL~\cite{psl}~(\textit{ECCV-24}) & d & \xmark &  &8.78 &3.30 &-      && 8.91 & 2.83 & - &&& 4.58 & 1.39 & -           \\

         OVON$^\dagger$~\cite{yokoyama2024hm3d_ovon}~(\textit{IROS-24})  & c & \xmark &  &8.18$^\dagger$& 5.24 &-      && 15.88$^\dagger$ & 11.35 & - &&&2.61$^\dagger$ & 1.29 & -         \\\midrule
         
         VFLM~\cite{yokoyama2023vlfm}~(\textit{ICRA-24})  & c & \mycheckmark &  &0.36& 0.28  &-      && 0.00 & 0.00 & - &&&0.00 & 0.00 & -         \\

         \textbf{\ourMethod}  (\textit{ours})& c & \mycheckmark &  &\textbf{7.42} & 2.92 & 1.67   &&   \textbf{14.38}& 7.99 & 1.36 &&& \textbf{6.67}&2.30 &1.13    \\\bottomrule                            
\end{tabular}}
\caption{\ourDataset{} is challenging. AIUTA, while being \textit{training-free}, achieves strong performance by outperforming trained policies (top rows) and significantly surpassing the zero-shot VLFM, across \textit{all} splits, through effective user interaction.
In contrast, policies trained on GOAT-Bench (denoted with \textbf{$^\dagger$}), the foundation of CoIN-Bench, fail to generalize to novel categories (Val Unseen).
We report the \texttt{SR} (main metric, in \textbf{bold} \textit{w.r.t} training free-methods), \texttt{SPL}, and the number of questions \texttt{NQ}. Input types: \textit{c} for object category, \textit{d} for its description.}
\label{table:main_results}
\end{table*}

\noindent\textbf{Evaluation protocol.} 
\ourDataset~supports evaluation with both \textit{real humans}, to assess the potential and limitation of genuine agent-human interactions, and simulated user-agent interactions, to enable extensive, reproducible and large-scale experiments.
Simulating agent-human interactions is challenging due to:~\textit{(i)}~ the agent's open-ended, template-free questions about any target attribute, making it impractical to predefine a comprehensive question-answer dataset, and \textit{(ii)} the huge question space in the simulated continuous environment~\cite{Savva_2019_ICCV_habitat}.
To address this, we propose to simulate user responses via a VLM with access to a high-resolution image of the target object ($1024\times1024$) at each episode. This setup is more effective than relying solely on instance descriptions~\cite{think_act_ask}, as the comprehensive visual coverage allows for diverse responses to the agent’s questions.

\noindent\textbf{Metrics.}
An episode is successful if the agent stops within $0.25$m of the target goal viewpoints. If not located, the exploration ends after $500$ actions.
Following~\cite{anderson2018evaluation, habitatchallenge2023}, we use: Success Rate, \texttt{SR}~($\uparrow$), our primary metric (in gray), and Success rate weighted by Path Length, \texttt{SPL}~($\uparrow$). 
Additionally, we introduce the \emph{average Number of Questions asked}, \texttt{NQ}~($\downarrow$) in successful episodes to measure the amount of user input.

\section{Experiments}
\label{sec:experiments}
We first benchmark \ourMethod{} against state-of-the-art (SOTA) methods~\cite{psl, goat_bench, yokoyama2023vlfm,yokoyama2024hm3d_ovon}~on \ourDataset{}, with simulated user-agent interactions, highlighting that \ourDataset{} present a challenging evaluation set for \textit{training-free} and training-based methods.
Next, we conduct an evaluation on a small validation set using both real human and simulated user-agent interactions, demonstrating that the simulation setup serves as a viable alternative to real human evaluation, enabling scalable and reproducible experiments. 
Finally, ablation studies validate \ourMethod{} design choices, and showcase the effectiveness of the Normalized-Entropy-based technique for estimating VLM uncertainty, outperforming recent baselines~\cite{zhao2024first, energy_based_OOD} on the \ourDatasetVQA{} dataset.

\noindent\textbf{Implementation Details.}
We use~\cite{liu2024llavanext} (LLaVA 1.6, Mistral 7B) as the VLM and GPT-4o~\cite{hurst2024gpt} as the LLM.
User interaction is limited to a maximum  of $4$ rounds. We empirically set $\tau = 0.75$ (Eq.~\ref{eq:abstein}), $\tau_{stop}=7$ and $\tau_{skip}=5$ as they yield the best result. In \suppmat{}, see \cref{supmat:prompts} for all prompts and \cref{supmat:computational_analysys} for AIUTA's computational analysis.

\noindent\textbf{Baselines.} We compare AIUTA against SOTA Instance Navigation and ObjectNav methods: the SenseAct-NN Monolithic Policy (\monolithic)~\cite{goat_bench}, PSL~\cite{psl}, OVON~\cite{yokoyama2024hm3d_ovon} and the zero-shot, training-free VLFM~\cite{yokoyama2023vlfm}.

To demonstrate the challenging nature of our dataset, we include two baselines, \monolithic~\cite{goat_bench} and OVON~\cite{yokoyama2024hm3d_ovon}, which are trained on GOAT-Bench. Again, note that the ``\textit{Seen}" splits contains categories seen during training (Sec.~\ref{sec:benchmark}).
PSL is trained on the ImageNav task and transferred on the language-driven Instance navigation task.
Notably, both \monolithic and PSL take a fully detailed description \textit{d} of the target instance as input, while OVON~\cite{yokoyama2024hm3d_ovon} takes the target category \textit{c}. Finally, VLFM operates
in a zero-shot, training-free manner, while taking category $c$ in input. All baselines are detailed in \suppmat{} \cref{supmat:baselines}. Tab.~\ref{table:main_results} summarizes the input types and training conditions on \ourDataset{}.

\noindent\textbf{Results with simulated user-agent interaction.}
As shown in Tab.~\ref{table:main_results}, training-based methods perform better on \texttt{Val Seen} and \texttt{Val Seen Synonyms} than on \texttt{Val Unseen}, highlighting their poor generalization to novel categories. 
This phenomenon is particularly pronounced on policies trained on GOAT-Bench (denoted with $^\dagger$), with performance dropping significantly—OVON's \sr decreases from a maximum of $15.88$ to $2.61$, and Monolithic's \sr drops from $13.09$ to $0.22$.
In contrast, \ourMethod{}, while being training-free, outperforms training-based methods on \texttt{Val Unseen}, with consistent and strong results in all the splits.
Interestingly, on the \texttt{Val Seen Synonyms}, \ourMethod{} is inferior to OVON, but outperforms PSL and Monolithich in \sr and \spl.
This is surprising, as PSL and Monolithic are training-based and operate with detailed instance descriptions. One possible explanation is that CLIP-based approaches is limited in encoding fine-grained instance description compared to category~\cite{goat_bench,cows}.
Moreover, compared to the results reported on GOAT-Bench, the lower \sr of the baselines, \eg Monolithic~\cite{goat_bench} on \ourDataset{}, highlights the introduced challenge of multi-instance ambiguity.

In particular, our closest competitor VLFM~\cite{yokoyama2023vlfm}, when using only the instance category as input, fails nearly all evaluation episodes, with almost $0\%$ \sr across all splits. 
This is expected, as the large amount of distractor objects (\cref{tab:episode_counts}) poses significant challenges for ObjectNav methods,  which lack instance-level discrimination capabilities. Further analysis of VLFM $0\%$ \sr can be found in \suppmat~\cref{supmat:distractors_success}.
In contrast, despite being built on top of VLFM and taking only the instance category, \ourMethod{} effectively gather additional information from the user to identify the correct instance, requiring minimal agent-user interaction (\texttt{NQ}$<2$ for all splits).
This results in a substantial improvement in \sr, achieving an outstanding $\sim14\times$ increase on \texttt{Val Seen Synonyms}, $\sim7\times$ on \texttt{Val Seen}, and approximately $\sim7\times$ on \texttt{Val Unseen}.
We illustrate the diversity of AIUTA-generated questions in \suppmat~Sec.~\ref{supmat:umap}.

\noindent\textbf{Validation with real human.}
To validate that simulated user-agent interactions yield credible results, we further conduct evaluation with real human on a small subset of \ourDataset{}. 
We randomly select $40$ episodes with \textit{detectable} target instances across all splits to minimize time and cognitive load. 
As a result, the \sr for this set are higher compared to those reported in \cref{table:main_results}.
We engage $20$ participants of varying ages and backgrounds, each evaluating two episodes.
Participants are provided an image depicting the target instance and interact with the agent via a chat-like interface (see \textbf{aiuta\_demo.mp4} in the supplementary materials).
They initiates the navigation via the fixed template ``\texttt{Find the <category>}'', and answer the agent's questions in natural language. 
More details about human evaluation in \suppmat{}~\cref{supmat:human_eval}.
The human results is compared against with simulated user-agent interactions in Tab.~\ref{table:human_exp}. We observe no significant differences in main metrics, confirming that the simulation setup is \textit{reliable for reproducible evaluation}.

\begin{table}[t!]
\centering
\resizebox{0.30\textwidth}{!}{
\begin{tabular}{ccsccc}
    \toprule
    \multirow{1}{*}{User type} & \multicolumn{4}{c}{\ourDataset{} subset}  \\ \cmidrule{3-5} &&\scriptsize\textbf{SR}~$\uparrow$ & \scriptsize\textbf{SPL}~$\uparrow$ &\scriptsize\textbf{NQ}~$\downarrow$ \\  \midrule 
    Simulated   && 42.50 & 15.48 & 1.10 \\
    Real Human  && 42.50 & 17.44 & 1.29\\ 
    \bottomrule
\end{tabular}}
\caption{Real human \textit{vs} simulated user-agent interaction.}
\label{table:human_exp}
\end{table}

\begin{table}
\centering
\resizebox{0.40\textwidth}{!}{
\begin{tabular}{ccc sccc}
    \toprule
    \multirow{2}{*}{Self-Questioner} & \multirow{2}{*}{Skip-Question} & \multicolumn{4}{c}{Ablation split} \\ 
                             \cmidrule{4-6} &   && \scriptsize\textbf{SR}~$\uparrow$ & \scriptsize\textbf{SPL}~$\uparrow$ &\scriptsize\textbf{NQ}~$\downarrow$   \\ \midrule
                                \xmark & \xmark &   & 9.21 & 5.86 & 3.57   \\                             
                                 \xmark & \mycheckmark &   & 8.55 & 4.84 & 2.69   \\
                                 \mycheckmark & \xmark &   & 9.87 & 6.5 & 4.6 \\ \midrule
                                 \mycheckmark & \mycheckmark &   & \textbf{14.47} & \textbf{7.22} &  \textbf{1.68}\\ \bottomrule
                                
\end{tabular}}
\caption{Ablation of components in \ourMethod{} on the \texttt{Train} split.}
\vspace{-0.5cm}
\label{table:ablations}
\end{table}

\noindent\textbf{\emph{Ablation I}: \ourMethod{} components.}
We introduce the Ablation split, derived from the largest GOAT-Bench \texttt{Train} split, following the procedure in Sec.~\ref{sec:benchmark}. We select GOAT-Bench \texttt{Train} as it covers more semantic categories. Since AIUTA is \textit{training-free}, validation remains fair.
Tab.~\ref{table:ablations} highlights the importance of the \textit{Self-Questioner} and \textit{Skip-Question} (within the Interaction Trigger).
Without both (row 1), \sr drops to $9.21\%$, with a high number of questions \texttt{NQ}.
Removing only the Self-Questioner (row 2) lowers the \sr, reducing \texttt{NQ}, as expected.
Enabling only the Self-Questioner (row 3) improves \sr to $9.87\%$, but keeps \texttt{NQ} high. With both components active (row 4), \sr peaks at $14.47\%$, and \texttt{NQ} drops to $1.68$, proving both effectiveness and efficiency.

\noindent\textbf{\emph{Ablation II}: VLM uncertainty estimation on \ourDatasetVQA{}.}
VLM uncertainty estimation is crucial for the Self-Questioner module, helping the agent to mitigate hallucinations and inaccuracies.
For validating these techniques, we introduce \ourDatasetVQA{}, a VQA dataset with $502$ questions and $102$ images from GOAT-Bench~\cite{goat_bench}.
Each question is answered by three annotators who choose from \{\texttt{Yes, No, I Don't Know}\}, allowing the agent to abstain when information is insufficient.
We compare our \textit{Normalized-Entropy}-based technique against three recent techniques:~MaxProb (selects the answer with the highest predicted probability); an energy score-based framework for out-of-distribution detection~\cite{energy_based_OOD}; and LP~\cite{zhao2024first}, a recent logistic regression model trained as a linear probe on the logits distribution of the first generated token. 
Tab.~\ref{table:vqa_model} reports the performance using the \textit{Effective Reliability} metric $\Phi_c$ proposed in~\cite{reliable_vqa}.
Our proposed technique achieves the best $\Phi_{c=1}$ score of $21.12$, demonstrating its effectiveness.
Further details in the \suppmat{} (Sec.~\ref{supmat:vqa_dataset}).

\begin{table}[t]
\centering
\resizebox{0.45\textwidth}{!}{
\begin{tabular}{ccs s}
    \toprule
    \multirow{1}{*}{VLM Model} & \multirow{1}{*}{Selection Function} & \multicolumn{1}{c}{$\Phi_{c=1}$} \\  \midrule
    \multirow{3}{*}{LLaVA llava-v1.6-mistral-7b-hf} & MaxProb &  15.94 \\
    & LP~\cite{zhao2024first} & 14.01 \\
    & Energy Score~\cite{energy_based_OOD} & 20.45 \\ \cmidrule{2-3}
    & \textbf{Normalized Entropy} (ours) & \textbf{21.12} \\ 
    \bottomrule
\end{tabular}}
\caption{Results of different selection functions and their corresponding \textit{Effective Reliability} rate $\Phi_{c=1}$ on the \ourDatasetVQA{} dataset.}
\label{table:vqa_model}
\end{table}

\noindent\textbf{\emph{Ablation III}: $\tau$.}
We analyze the sensitivity of the threshold ($\tau$ in Eq.~\ref{eq:abstein}) for our \textit{Normalized-Entropy}-based technique and second-best performing Energy Score~\cite{energy_based_OOD}. 
We subsample the datasets to $50\%$, $70\%$, and $100\%$ of its original size. 
For each subsampled dataset, we find the optimal threshold $\tau^*$ and evaluate its sensitivity by testing $\Phi_{c=1}$ on $30$ alternative thresholds around $\tau^*$, normalizing it between $0$ and $1$. 
As shown in Fig.~\ref{fig:sensitivity}, our technique has a smaller interquartile range and a tighter distribution of $\Phi_{c=1}$, while~\cite{energy_based_OOD} exhibits a greater degradation from $\tau^*$, which worsens as the dataset size decreases.
This proves that our technique is more robust in data-scarce situations, and is less sensitive to small variations in $\tau$.
Moreover,~\cite{energy_based_OOD} depends on logits, thus being unbounded. On the contrary, our uncertainty is normalized, \ie $u \in [0, 1]$, making optimal $\tau$ selection more efficient. 
\begin{figure}
    \centering
    \includegraphics[width=0.75\linewidth]{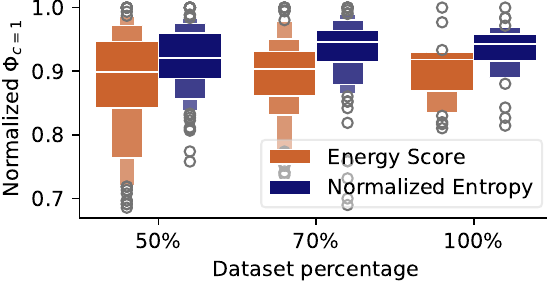}
    \caption{$\tau$ sensitivity results. For each method, $30$ new $\tau$ values are sampled symmetrically around the optimal threshold $\tau^*$. 
    The $x$-axis shows the set size as a percentage of the original \ourDatasetVQA{} dataset size, while the $y$-axis displays the normalized ER $\Phi_{c=1}$.}
    \label{fig:sensitivity}
    \vspace{-0.6cm}
\end{figure}

\section{Conclusion}
\label{sec:conlusion_and_limitation}
We introduced the \ourTask{} task, where the agent collaborates with the user during navigation to resolve uncertainties about the target instance.
Trough extensive experiments, we show that existing trained method fails to generalize to unseen categories, while our training-free AIUTA, using a novel self-dialogue mechanism and uncertainty estimation, achieves strong performance across all validation splits.
Moreover, our simulated user-agent interaction is in line with human evaluation, enabling scalable and reproducible experiments. 

\ourMethod{} relies on current LLMs, where larger models improve performance but at a high inference cost, limiting real-time on-board processing. Future works will investigate model optimization for embodied deployment, and extending the interaction scope to action instructions.

{
    \small
    \bibliographystyle{ieeenat_fullname}
    \bibliography{main}
}

\clearpage
\setcounter{page}{1}
\setcounter{figure}{0}
\setcounter{equation}{0}
\setcounter{section}{8}
\renewcommand{\thesection}{\AlphAlph{\value{section}-8}}
\maketitlesupplementary
In this supplementary material, we first provide additional details regarding the \ourDataset{} dataset (\underline{Sec.~\ref{supmat:ourdataset}}), including an overview of the GOAT-Bench dataset on which \ourDataset{} is based, as well as statistics and examples of target instances within \ourDataset{}. 
Additionally, in \underline{Sec.~\ref{supmat:task_setup}}, we provide a visualization of the evaluation setup, clarifying the role of the user responses in our evaluation, whether from real human or simulation.
Next, we elaborate on the implementation details for baseline comparisons in \underline{Sec.~\ref{supmat:baselines}}, and present a computational analysis of AIUTA in 
\underline{Sec.~\ref{supmat:computational_analysys}}.
Further details on the real human experiments are provided in \underline{Sec.~\ref{supmat:human_eval}}.
Then, in \underline{Sec.~\ref{supmat:distractors_success}} we investigate the low performance of the original VLFM, and 
in \underline{Sec.~\ref{supmat:umap}}, we provide a UMAP visualization of the questions generated by the agent.
Next, in \underline{Sec.~\ref{supmat:vqa_dataset}}, we detail the evaluation conducted on \ourDatasetVQA{}  including the dataset creation, evaluation metric, and state-of-the-art baselines used for comparison.
Finally, we include all the prompts in \underline{Sec.~\ref{supmat:prompts}} and the full algorithm of \ourMethod{} in \underline{Sec.~\ref{supmat:algorithm}}. 

For a demonstration of \ourMethod{} in action, engaging with a real human through natural language dialogues to collaboratively localize a target instance, please refer to the accompanying video ({\color{blue} \textbf{aiuta\_demo.mp4}}) provided in the supplementary material.

\section{Additional details of \ourDataset}
\label{supmat:ourdataset}
\subsection{CoIN-Bench}
\begin{figure*}[t!]
\centering
    \includegraphics[width=1\linewidth]{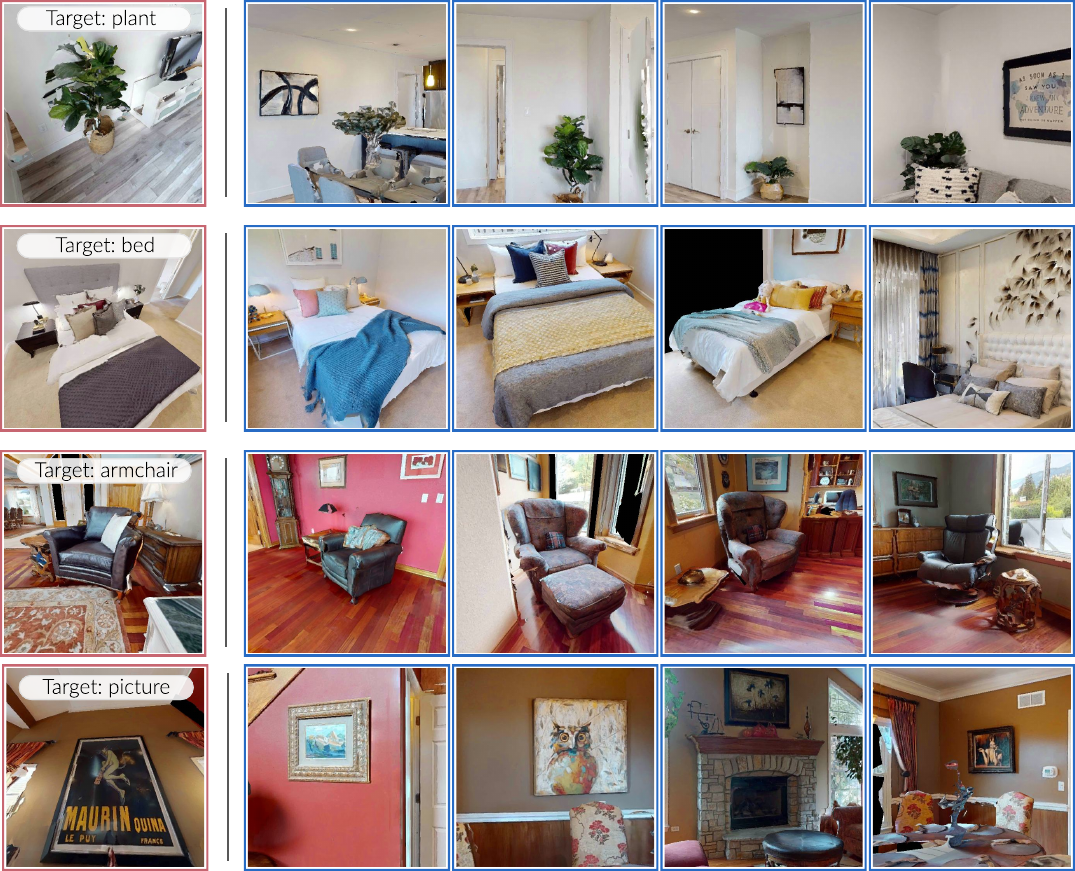}
  \caption{\ourDataset{} can be very challenging when only given the instance category to the agent. We highlight the target instance with red borders, while the distractor instances that exist in the same scene are marked with blue borders.}
\label{fig:sup:coin_bench2}
\end{figure*}

\noindent\textbf{Instance examples.}
The \ourDataset{} benchmark poses a significant challenge, since multiple distractor objects are present among each target. 
To illustrate this, Fig.~\ref{fig:sup:coin_bench2} provides examples where the target instance is highlighted with red borders, while distractors in the same scene are marked with blue borders. As demonstrated, agent-user collaboration is crucial to gather the necessary details for uniquely identifying the target instance among other visually similar objects of the same category, such as the armchair or the plant.

\noindent\textbf{Dataset statistics.}
We provide additional statistics for the \ourDataset{} dataset.
In Fig.~\ref{fig:sup:plot_statistic_dataset} we show the shortest path statistics for the \ourDataset{} dataset. In particular, the euclidean and geodesic distance for all the split, as well as the number of distractors.
Next, Fig.~\ref{fig:sup:bar_plot_cat} illustrates the distribution of instance categories across different splits. These splits are ordered by dataset size, from the largest at the top (\texttt{Val Seen}) to the smallest at the bottom (\texttt{Val Seen Synonyms}). 
The number of distinct categories decreases as the dataset size reduces. The \texttt{Val Seen} split, being the largest, also contains the highest number of distinct categories, with ``cabinet'', ``bed'',  and ``table'' being the top 3 common categories. \texttt{Val Seen Synonyms}, being the smallest, only contains 3 categories.

\begin{figure*}[h!]
    \centering
    \includegraphics[width=1\textwidth]{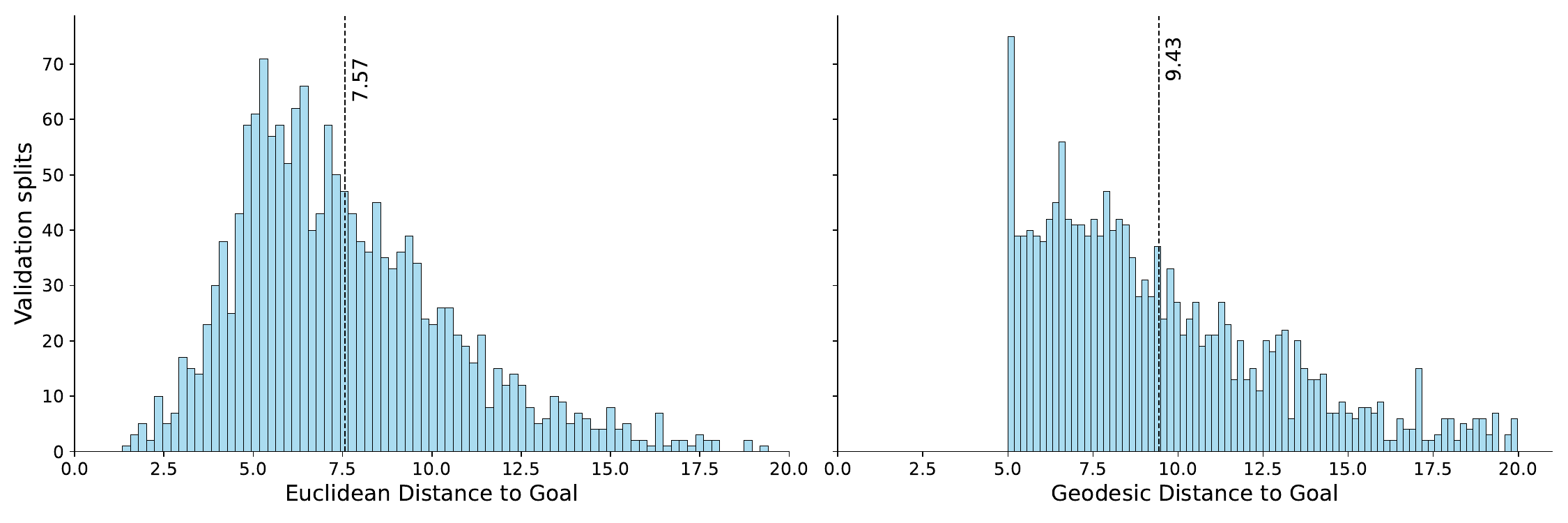}
    \caption{Distribution of the path length to the goal, both in Euclidean and Geodesic term.}
    \label{fig:sup:plot_statistic_dataset}
\end{figure*}

\begin{figure*}[t!]
    \centering
    \includegraphics[width=1\textwidth]{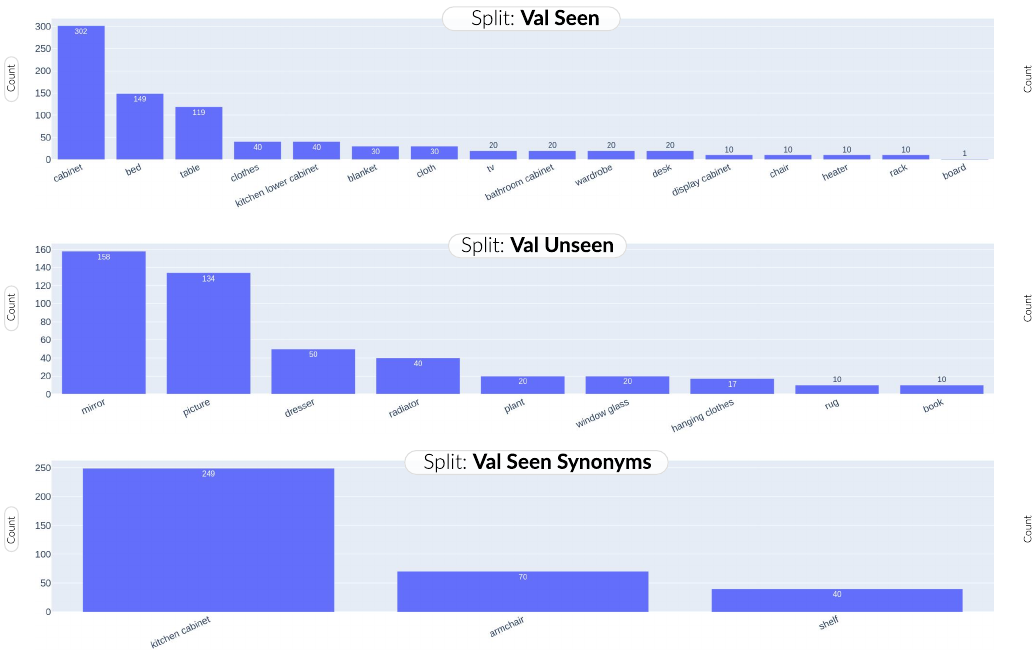}
    \caption{We show the distribution of categories, categorized for each evaluation split.}
    \label{fig:sup:bar_plot_cat}
\end{figure*}

\begin{figure}[th!]
    \centering
    \includegraphics[width=0.45\textwidth]{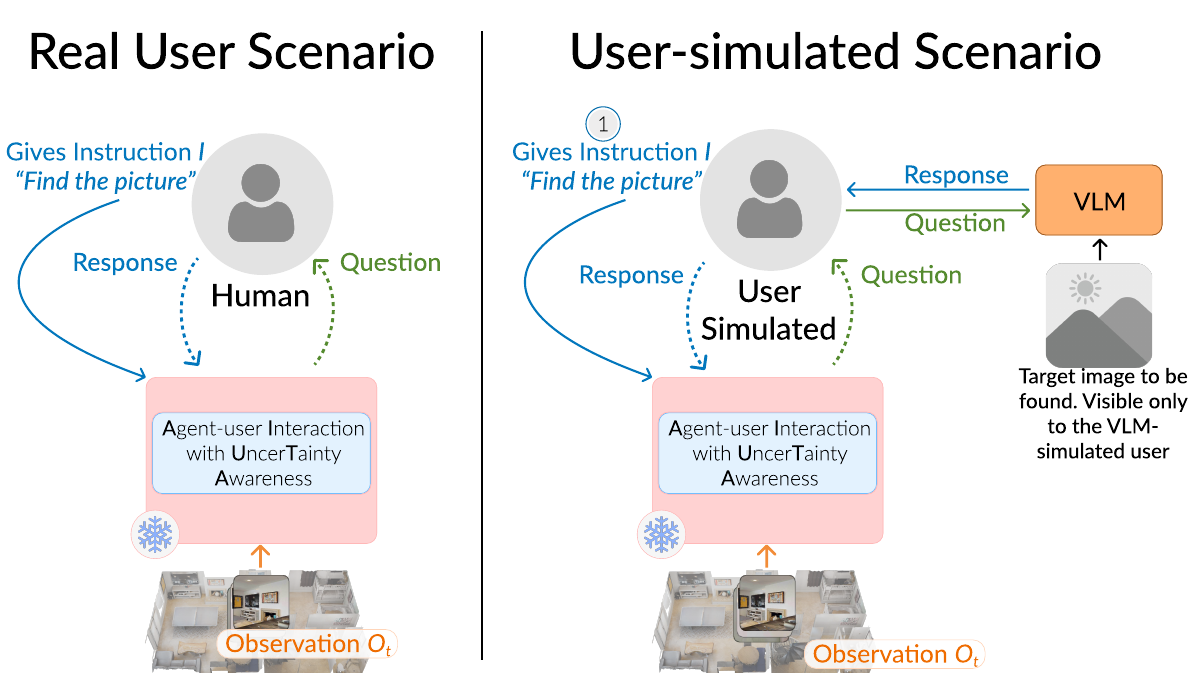}
    \caption{\ourDataset{} evaluation setup. (Left) Real human responding to the agent's question. (Right) Simulated user-agent interactions, where the user responses are provided by a VLM with access to a high-resolution target instance image for scalable and reproducible experimentation.}
    \label{fig:sup:evaluatio_setup}
\end{figure}

\subsection{GOAT-Bench}
\label{supmat:goat_bench}
\noindent\textbf{Dataset.}
GOAT-Bench provides agents with a sequence of targets specified either by category name $c$ (using episodes from~\cite{yokoyama2024hm3d_ovon}), language description $d$, or image in an open vocabulary fashion, using the HM3DSem~\cite{hm3d_sem} scene datasets and Habitat simulator~\cite{Savva_2019_ICCV_habitat}. 
Natural-language descriptions $d$ are created with an automatic pipeline by leveraging ground-truth semantic and spatial information from simulator~\cite{Savva_2019_ICCV_habitat} along with capabilities of VLMs and LLMs. 
Specifically, for each object-goal instance, a viewpoint image is sampled to maximize frame coverage.
From this sampled image, the names and 2D bounding box coordinates of visible objects are extracted.
Then, spatial information is extracted with the BLIP-2~\cite{blip_2} model, while ChatGPT-3.5 is prompted to output the final language description.

\noindent\textbf{Splits.} 
GOAT-Bench baselines are trained on \texttt{Train} split, and evaluated on validations splits.
Notably, the evaluation splits are divided into \texttt{Val Seen} (\ie, object categories seen during training), \texttt{Val Seen Synonyms} (\ie, object categories that are synonyms to those seen during training) and \texttt{Val Unseen} (\ie, novel object categories).

\section{Evaluation Setup} \label{supmat:task_setup}
In Fig.~\ref{fig:sup:evaluatio_setup}, we show the two evaluation setups, highlighting their differences between the human user and the simulated user-agent interactions. Fig.~\ref{fig:sup:evaluatio_setup} (Left) shows how a human user answers the agent's queries based on their knowledge of the target instance. 
However, relying on human responses for large-scale evaluations is impractical due to variability, scalability constraints and large cost. To address this, we introduce a simulated user-agent interaction setup as in Fig.~\ref{fig:sup:evaluatio_setup} (Right). The user responses are simulated via a VLM with access to the high-resolution target instance image, which is never available to the agent.
With the visual coverage of the target instance, the simulated user responses can support the diverse open-ended, template-free questions from the agent, about any attribute of the target instance.
This is more desired than previous work~\cite{think_act_ask} whose simulation setup leverages an LLM with access to the instance description, particularly when the instance description misses critical fine details that the agent deems important to know.
For instance, in the case of the picture in Fig.~\ref{fig:teaser} of the main paper, the instance description may not mention ``the person is shirtless", but this detail is critical for the agent to eventually disambiguate the target instance from distractors.

\section{Baselines} \label{supmat:baselines}
In this section, we provide a description of the different baselines for Instance Navigation and Object Navigation used throughout the paper: VLFM (Sec.~\ref{supmat:baseline:vlfm}), \monolithic (Sec.~\ref{supmat:baseline:goat_mono}), PSL (Sec.~\ref{supmat:baseline:psl}), and OVON (Sec.~\ref{supmat:baseline:ovon}).

\subsection{VLFM}
\label{supmat:baseline:vlfm}
VLFM~\cite{yokoyama2023vlfm} is a zero-shot state-of-the-art object-goal navigation policy that does not require model training, pre-built maps, or prior knowledge about the environment. 
The core of the approach involves two maps: a frontier map (see in Fig.~\ref{fig:sup:vlfm_map}~(a)) and a value map (see Fig.~\ref{fig:sup:vlfm_map}~(b)).

\noindent\textbf{Frontier map.}
The frontier map is a top-down 2D map built from depth and odometry information. 
The explored area within the map is updated based on the robot's location, heading, and obstacles by reconstructing the environment into a point cloud with the depth images, and then projecting them onto a 2D grid. 
The role of the frontier map is to identify each boundary separating the explored and unexplored areas, thus identifying the frontiers (see the blue dots in Fig.~\ref{fig:sup:vlfm_map}~(a)).

\noindent\textbf{Value map.}
The value map is a 2D map similar to the frontier map. For each point within the explored area, a value is assigned by quantifying its relevance in locating the target object (see Fig.~\ref{fig:sup:vlfm_map}~(b)).
At each timestep, frontiers are extracted from the frontier map, and the frontier with the highest value on the value map is selected as the next goal for exploration.
To efficiently guide the navigation, VLFM projects the cosine similarity between the current visual observation and a textual prompt (\eg, \textit{“Find the picture”}) onto the value map. This similarity is computed using the BLIP-2 model~\cite{blip_2}, which achieves state-of-the-art performance in image-to-text retrieval.
To verify whether a target instance is present in the current observation, VLFM employs Grounding-DINO~\cite{grounding_dino}, an open-vocabulary object detector. Once a candidate target is detected, Mobile-SAM~\cite{mobile_sam} refines the detection by segmenting the object’s contour within the bounding box. The segmented contour is paired with depth information to determine the closest point on the object relative to the agent’s position. This point serves as a waypoint for the agent to navigate toward the object.

At each timestep, the action $a_t$ is selected using a PointGoal navigation (PointNav) policy~\cite{anderson2018evaluation}, which can navigate to either a frontier or a waypoint, depending on the context.
\label{supmat:vlfm}
\begin{figure}[t!]
    \centering
    \includegraphics[width=0.35\textwidth]{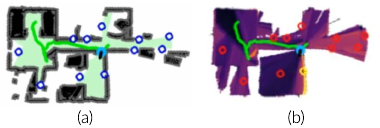}
    \caption{(a) Frontier map and (b) value map constructed by VLFM~\cite{yokoyama2023vlfm}. The blue dots in (a) (as well as the red dots in (b)) are the identified frontiers.}
    \label{fig:sup:vlfm_map}
\end{figure}

\subsection{\monolithic}
\label{supmat:baseline:goat_mono}
The \monolithic (SenseAct-NN Monolithic Policy) is a single, end-to-end reinforcement learning (RL) policy designed for multimodal tasks, leveraging implicit memory and goal encoding proposed in~\cite{goat_bench}.
RGB observations are encoded using a frozen CLIP~\cite{clip} ResNet50 encoder.
Additionally, the agent integrates GPS and compass inputs, representing location $(\Delta x, \Delta y, \Delta z)$
and orientation ($\Delta \theta$). These inputs are embedded into 32-dimensional vectors using a encoder with fully connected layers.
To model multimodal inputs, a $ 1024$-dimensional goal embedding is derived using a frozen CLIP image or CLIP text encoder, depending on the subtask modality (object, image, or language). All input features—image, location, orientation, and goal embedding—are concatenated into an observation embedding, which is processed through a two-layer, 512-dimensional GRU. At each timestep, the GRU predicts a distribution over a set of actions based on the current observation and the hidden state.
The policy is trained using 4$\times$A40 GPUs for approximately 500 million steps.

\subsection{PSL}
\label{supmat:baseline:psl}
PSL~\cite{psl} is a zero-shot policy for instance navigation, which is pre-trained on the ImageNav task and transferred to achieve object goal navigation without using object annotations for training.

Built on top of ZSON~\cite{zson}, observations are processed by a learned ResNet50 encoder and a frozen CLIP encoder obtaining, respectively, observation embeddings and semantic-level embeddings.
To encode the goal modality, an additional frozen CLIP encoder is used, obtaining goal embedding.
The goal and the semantic-level embeddings are additionally processed by a semantic perception module, which reduces dimension condensing critical information, emphasizing the reasoning of the semantics differences in the goal and observation.
Based on condensed embeddings and observation embeddings, the authors trained a navigation policy using reinforcement learning.
Specifically, the PSL agent is trained for 1G steps following ZSON~\cite{zson}, on 16 Nvidia RTX-3090 GPUs.

\subsection{OVON}
\label{supmat:baseline:ovon}
OVON \citep{yokoyama2024hm3d_ovon} is a transformer-based policy designed for the open-vocabulary object navigation task. At each timestep $t$, it constructs a $1568$-dimensional latent observation $o_t$ of the current navigation state by concatenating the encodings of the current image $I_t$, object description $c$, and previous action $a_{t-1}$, using two SigLIP encoders~\cite{siglip} and an action embedding layer. 
This latent observation is then passed to a $4$-layer decoder-only transformer~\cite{transformer}, which, along with the previous $100$ observations, outputs a feature vector.
This vector is then used to produce a categorical distribution over the action space via a simple linear layer.
The policy is trained on their proposed HM3D-OVON dataset using various methods, such as RL and BC, for $150$M to $300$M steps, across $16$ environments on $8\times$ TITAN Xp, resulting in a total of $128$ environments. Note that the categories seen in training in HM3D-OVON overlap with that of GOAT-Bench.

\section{Computational analysis}
\label{supmat:computational_analysys}
In the following, we report the average inference time of AIUTA over 20 episodes, using an NVIDIA $4090$ GPU, following the steps outlined in the algorithm in Sec.~\ref{supmat:algo:self_q}: \textit{Step $1$}, Detailed Detection Description: 11.3s; \textit{Step 2,} Perception Uncertainty Estimation: 8.29s; Step 3 + Interaction Trigger: 6.13s.
We would like to emphasize that our code was not optimized for speed, as it is out of scope of our study- we did not apply model compilation (\eg, \texttt{torch.compile}) and quantization, leaving room for further efficiency improvements.
Moreover, in AIUTA, we identify the primary bottleneck is the LLM call.
As discussed in the \textit{Conclusion} (Sec.~\ref{sec:conlusion_and_limitation}) and in~\cite{chiang2024mobility}, reducing model dimensionality while maintaining similar reasoning performance is a necessity and a promising direction for future work.
Additionally, our inference time is in line with other works~\cite{chiang2024mobility}. 
Finally, the emerging field of Language Processing Units (LPUs) offers potential solutions, promising near-instant inference, high affordability, and energy efficiency at scale~\cite{groq2024}.

\section{Evaluation with real human}
\label{supmat:human_eval}
To demonstrate the reliability and reproducibility of our simulated setup, we run a human study comparing the performance of \ourMethod{} when user responses are provided by:~\textit{(i)}~\textit{real human} and \textit{(ii)}~\textit{simulations} (Fig. \ref{fig:sup:evaluatio_setup}). 
A total of 20 volunteers participated in the study (12 males and 8 females), with ages ranging from 20 to 40 years. 
All participants have backgrounds in electronic engineering, computer science, or other relevant fields, minimizing expertise barriers to conducting the experiments.
At the start of each episode, participants are given an image depicting the final target instance, which remains accessible throughout the experiment.
Again, note that this image is never seen by the agent.
This setup simulates a real-world scenario where a human has a reference image in mind, enabling them to answer questions correctly.
The human user then initiates the navigation by sending the initial instruction to the agent (using the fixed template ``\texttt{Find the <category>}'') via a chat-like User Interface (UI) that we have developed for the evaluation (as demonstrated in the supplementary video, \textbf{aiuta\_demo.mp4}). 
Next, the human user is encouraged to respond to the questions posed by \ourMethod{} in natural language and to truthfully reflect the facts about the target instance. 
For this evaluation, we have selected 40 episodes across \ourDataset{} dataset, randomly distributed among participants, with each conducting two evaluations.
When compared with the simulated setting, we found no statistical differences in terms of results, showing that our simulated-based evaluation is reliable and reproducible.

\section{VLFM results}
In this section, we investigate the low \texttt{SR} results of VLFM in Tab. \ref{table:main_results} of the main paper. To better understand this behavior, we introduce an additional metric, \texttt{Distractor Success}, which mirrors the success rate but considers an episode successful if the agent stops at a distractor object instead of the target. As we can see in Tab.~\ref{tab:distractors_success}, VLFM successfully locates the correct category instance (\texttt{Distractor Success)} but struggles to discern its attributes and differentiate between instances (low \texttt{SR)}. Furthermore, this analysis highlights that the presence of sufficient distractors is well realized with our dataset construction procedure.
\label{supmat:distractors_success}

\begin{table}[h!]
    \centering
    \resizebox{1\columnwidth}{!}{%
    \begin{tabular}{l|cccc}
        \toprule
        \textbf{Statistics} & \textbf{Val Seen} & \textbf{Val Seen Synonyms} & \textbf{Val Unseen}  \\
        \midrule
        \texttt{SR} & 0.36 & 0.00 & 0.00 \\ \midrule
        \texttt{Distractor Success}  & 3.37 & 0.84 & 4.58 \\
        \bottomrule
        
    \end{tabular}
    }
    \caption{\texttt{SR} and \texttt{Distractor Success} comparison.}
    \label{tab:distractors_success}
\end{table}

\section{UMAP visualization}
\label{supmat:umap}
To illustrate the diversity of questions to the user generated by AIUTA, we collect $414$ question samples made by the agent, compute embedding using Sentece-Bert~\cite{sentence_bert} and visualize them using UMAP~\cite{umap} for dimensionality reduction. The results, shown in Fig.~\ref{fig:sup:umap}, demonstrate that AIUTA generates questions covering a wide range of attributes, such as color, material, style, and spatial arrangement.
\begin{figure}[t!]
    \centering
    \includegraphics[width=0.48\textwidth]{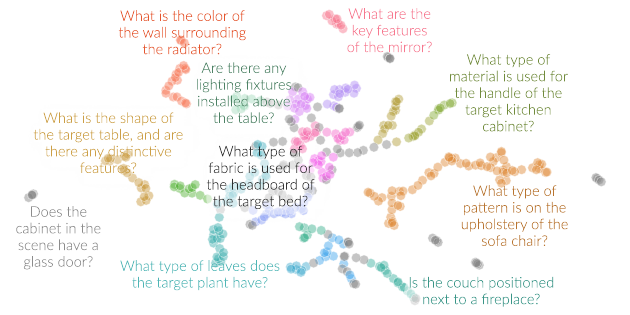}
    \caption{AIUTA generates questions covering a wide range of attributes, such as color, material, style, and spatial arrangement.}
    \label{fig:sup:umap}
\end{figure}

\section{\ourDatasetVQA{} dataset}
\label{supmat:vqa_dataset}
\begin{figure}[t!]
    \centering
    \includegraphics[width=0.45\textwidth]{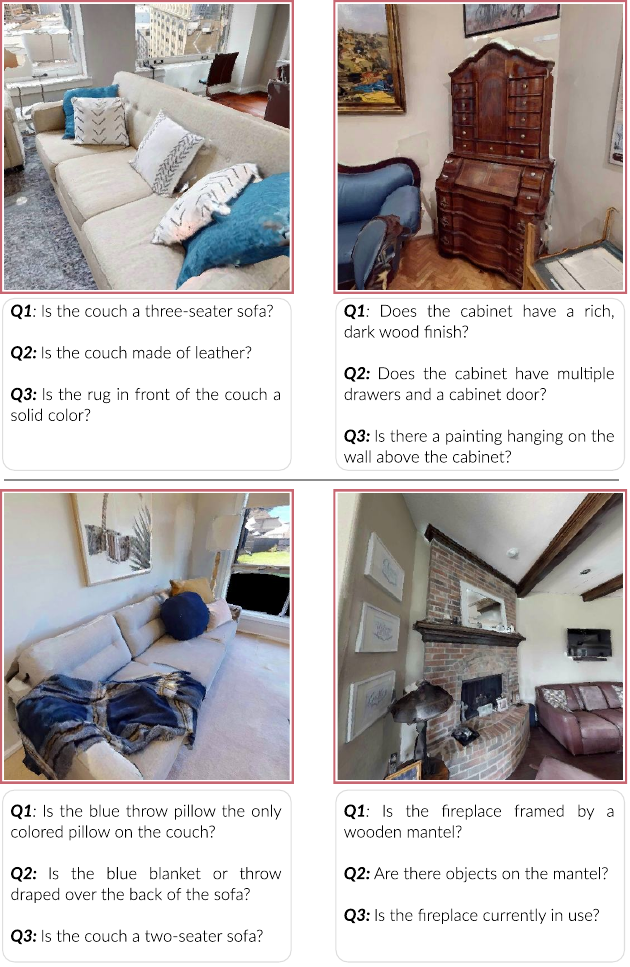}
    \caption{Examples from \ourDatasetVQA{}, showing images and the questions generated by the LLM.}
    \label{fig:sup:IDKVQA}
\end{figure}
\subsection{Dataset}
An essential feature of the \textit{Self-Questioner} is its ability to generate self-questions aimed at extracting additional attributes from the observation $O_t$ and assessing the uncertainty of the VLM. However, there exists no dataset in the such context for us to understand if how reliable a technique is for the VLM uncertainty estimation.

For this purpose, we introduce \ourDatasetVQA{}, a dataset specifically designed and annotated for visual question answering using the agent's observations during navigation, where the answer includes not only \texttt{Yes} and \texttt{No}, but also \texttt{I don't know}.
Specifically, we sample $102$ images from the training split of GOAT-Bench.
Then, for each image, we leverage the \textit{Self-Questioner} pipeline to generate a set of questions. Each question is annotated by three annotators, that can pick one answer from the set $\{\texttt{Yes},~\texttt{No}, \texttt{I don’t know}\}$.
Fig.~\ref{fig:sup:IDKVQA} illustrates sample images and their questions generated by the \textit{Self-Questioner} module.

\subsection{VLM uncertainty estimation on \ourDatasetVQA{}.}
In this section, we present a detailed analysis of VLM uncertainty estimation on \ourDatasetVQA{}, focusing on the evaluation metric and baseline methods.

\noindent\textbf{Metric.}
We evaluate the performance using the \textit{Effective Reliability} metric $\Phi_c$ proposed in~\cite{reliable_vqa}.
This metric captures the trade-off between risk and coverage in a VQA model by assigning a reward to questions that are answered correctly, a penalty $c$ to questions answered incorrectly, and a zero reward to the model abstaining. Formally:
\[
\Phi_c (x)= 
\begin{cases}
\text{Acc($x$)}, & g(x) = 1 \, \text{and} \, \text{Acc($x$)} > 0 \\
-c &  g(x) = 1 \, \text{and} \, \text{Acc($x$)} = 0 \\
0, & g(x) = 0
\end{cases}
\]

Here, $x=(i, q) \in \mathcal{X}$ is the input pair where $i$ is the image and $q$ is the question. The function $g(x)$ is equal to $1$ if the model is answering and $0$ if it abstains.
The parameter $c$ denotes the cost for an incorrect answer, and the VQA accuracy  \text{Acc} is:
\[
    \text{Acc}(f(x, y)) = \text{min}\left(\frac{ \text{\# annotations that match } f(x)}{3}, 1\right)
\]
where the function $f: \mathcal{X} \rightarrow V$ output a response $r \in \mathcal{R}$ for each input pair $x$.

\noindent\textbf{Baselines.}
We evaluate our proposed \textit{Normalized Entropy} against three baseline methods:

~\textit{(i)}~MaxProb, which selects the response $r$ with the highest predicted probability from the VLM, given image $i$ and question $q$. Formally, $r = \text{VLM}(i, q)$. It does not incorporate uncertainty estimation.

~\textit{(ii)} LP~\cite{zhao2024first}, a recently proposed Logistic Regression model trained as a linear probe on the logits distribution of the first generated token. 
The model is trained on the \textit{Answerable}/\textit{Unanswerable} classification task using the VizWiz VQA dataset~\cite{Gurari_2018}, which includes $23,954$ images for training.
When applied to \ourDatasetVQA{}, the logistic regression model first predicts whether the question $q$ is \textit{Answerable} or \textit{Unanswerable}. If the question is deemed answerable, the response $r$ with the highest probability is selected among $\{\texttt{Yes},~\texttt{No}\}$; otherwise, the response \texttt{I don't know} is returned.

~\textit{(iii)} Energy score, an energy-based framework for out-of-distribution (OOD) detection~\cite{energy_based_OOD}. Following the implementation in~\cite{energy_based_OOD}, an energy score is computed to identify whether the given question-image pair is OOD. If the pair is classified as OOD, the response \texttt{I don't know} is returned; otherwise, the response with the highest probability is selected among $\{\texttt{Yes},~\texttt{No}\}$.

Finally, for our proposed \textit{Normalized Entropy} estimation, we link the abstention function $g(x)$ (\ie, determining whether the model abstains from answering) to Eq.~\ref{eq:abstein} in the main paper.
Specifically, $g(x) = 1$ if the Normalized Entropy classifies the model as \textit{certain}, and $g(x) = 0$ otherwise.  Then, if the model is deemed certain, we return the most probable answer $\{\texttt{Yes},~\texttt{No}\}$; otherwise, the response \texttt{I don't know} is selected.

\subsection{Sensitivity analysis of the threshold $\tau$ }

This section provides additional details about how small variations of the threshold parameter $\tau$ affect both our Normalized Entropy technique (Eq.~\ref{eq:abstein}%
) and the Energy Score~\cite{energy_based_OOD}, with respect to the target metric $\Phi_{\text{c}=1}$.

To conduct this analysis, we perform an ablation study on datasets of varying sizes, obtained by randomly sub-sampling \ourDataset{}.
Specifically, we create five sets containing $50\%$ of the question-answer pairs from \ourDataset{}, five sets comprising $70\%$ of the question-answer pairs, and also use the full dataset ($100\%$) for a total of $11$ datasets.

For each dataset, we identify the optimal threshold $\tau^*$ for each method through an exhaustive search over predefined ranges, resulting in 22 optimal thresholds (11 per method)

 Around each $\tau^*$, we define a neighborhood $\boldsymbol{\tau}$ comprising $30$ new thresholds $\tau$ sampled symmetrically around it. %
Our goal is to analyze how $\Phi_{\text{c}=1}$ changes across these neighborhoods: if the values are spread out, it means that the method is very sensitive to small changes of $\tau$ near the optimal value, whereas if they are more tightly distributed it means that it is more robust.

Therefore, for each method and related neighborhood $\boldsymbol{\tau}$, we compute 30 $\Phi_{\text{c}}$ values, one for each $\tau \in \boldsymbol{\tau}$, and \textit{normalize} them to the range$ [0, 1]$ by dividing each value by the best $\Phi_{\text{c}=1}$ found in $\boldsymbol{\tau}$. We do so to measure only the distribution of the $\Phi_{\text{c}=1}$ values, not their absolute values, and to help the comparison across datasets of the same size (otherwise, due to chance, they could have distributions of different values).
Finally, we aggregate all these normalized $\Phi_{\text{c}=1}$ scores across dataset size, resulting in Fig.\ref{fig:sensitivity} (main paper).

From the figure, we can see that our technique has smaller interquartile ranges and tighter distributions of $\Phi_{\text{c}=1}$, while the Energy Score~\cite{energy_based_OOD} exhibits larger tails, indicating more variance. Moreover, our method shows distributions more biased toward higher values (which would indicate smaller degradation \textit{w.r.t.} the best $\Phi_{\text{c}=1}$) than those of the Energy Score, and this gap increases as the dataset size decreases. 
This shows that our technique is generally more robust, especially in data-scarce situations, and less sensitive to small variations in $\tau$.

\section{Prompts}
\label{supmat:prompts}
\subsection{$P_{init}$ - Initial Description}
\begin{lstlisting}[language=Python]
P_init = """Describe the {target_object} in the provided image."""
\end{lstlisting}

\subsection{$P_{details}$ - Gather Additional Information}
\label{supmat:additional_information}
\begin{lstlisting}[language=Python]
P_details = """You are an intelligent embodied agent equipped with an RGB sensor, an object detector, and a Visual Question Answering (VQA) model. 
Your task is to explore an indoor environment to find a specific target {target_object}.
The detector has identified a {target_object}. The VQA model has provided the following description of the scene:

<START_OF_DESCRIPTION>
{distractor_object_description}
<END_OF_DESCRIPTION>

Based on your past interactions with the user, you know the following facts about the target picture: 
<START_TARGET_PICTURE_FACTS> 
{facts_about_the_target_picture} 
<END_TARGET_PICTURE_FACTS> 

Your task is to:
- ask more question to the VQA model on the detected {target_object} to maximize information gain.

Ensure your output follows the following format:

YAML_START # must be present to get the information back
attributes_of_the_image:
    <attribute name>: "<attribute value>" # summarize all the known attributes from the description, enclosed in " "
questions:
        <question_number>: "<question content>"
YAML_END # must be present to get the information back
      
Provide your reasoning step-by-step, after the YAML_END tag."""
\end{lstlisting}

\subsection{$P_{check}$ - Check detection with LVML}
\label{supmat:check_detection}

\begin{lstlisting}[language=Python]
 P_check = """Does the image contain a {target_object}? Answer with Yes, No or ?=I don't know."""
\end{lstlisting}

\subsection{$P_{self question}$ - Extract attributes and generate Self-Questions}
\label{supmat:self_questions)}
\begin{lstlisting}[language=Python]
P_ATTRIBUTES_AND_SELF_QUESTIONS = """
You are an intelligent embodied agent equipped with an RGB sensor, an object detector, and a Visual Question Answering (VQA) model. Your task is to explore an indoor environment to find a specific target {target_object}.
The detector has identified a {target_object}. The VQA model has provided the following description of the scene:

<START_OF_DESCRIPTION>
{distractor_object_description}
<END_OF_DESCRIPTION>

Based on your past interactions with the user, you know the following facts about the target picture: <START_TARGET_PICTURE_FACTS> {facts_about_the_target_picture} <END_TARGET_PICTURE_FACTS> 

Assume that the detected image description contains hallucinations. Your goal is to verify every attribute of the detected {target_object} description through questions. Formally:
- Detect possible hallucinations in the VQA model's description
- Get more information about the detected object.
Every question should be in this format: "<question content>? You must answer only with Yes, No, or ?=I don't know." This allows us to assess the likelihood of the answers.


Ensure your output follows the following format:
YAML_START # must be present to get the information back
attributes_of_the_image:
    <attribute name>: "<attribute value>" # summarize all the known attributes from the description, enclosed in " "

questions_for_detected_object: # question for the detected object, if any
    <Question number>:  "<question>? You must answer only with Yes, No, or ?=I don't know."
reasoning_for_detected_object:
    <Question number>: <reasoning>
YAML_END # must be present to get the information back

Provide your reasoning step-by-step, after the YAML_END tag."""
\end{lstlisting}

\subsection{$P_{refined}$ - Refined image description}
\label{sup:mat:refined_image_desc}
\begin{lstlisting}[language=Python]
P_refined = """
You are an intelligent embodied agent equipped with an RGB sensor, an object detector, and a Visual Question Answering (VQA) model. 
Your task is to refine an image description based on certainty estimates and user interactions.

Scenario:
The detector has identified a scene with a {target_object}. The VQA model provided this initial scene description:

<START_OF_DESCRIPTION>
{distractor_object_description}
<END_OF_DESCRIPTION>


Questions asked and responses:
<START_QUESTION_AND_RESPONSES>
{list_questions_answers_uncertainty_labels}
<END_QUESTION_AND_RESPONSES>

Task:
Using the questions/answer pairs with uncertainty labels, refine the image description. 
Since we have to find a {target_object}, put enphasis on it. Do not include in the description information that is labeled as uncertain.

Ensure your response follows the format below:
YAML_START # must be present to get the information back
attributes_of_the_image:
    <attribute name>: "<attribute value>" # summarize all the known attributes from the description, enclosed in " "
image_description_refined: <insert refined description>  # Ensure that the string does not contain a newline (\n) after the tag image_description_refined:
YAML_END # must be present to get the information back
      
Provide your reasoning step-by-step, after the YAML_END tag."""
\end{lstlisting}

\subsection{$P_{score}$ - Alignment score}
\label{supmat:alignment_score}
\begin{lstlisting}[language=Python]
P_score = """
You are an intelligent agent equipped with an RGB sensor, object detector, and Visual Question Answering (VQA) model.
Your goal is to identify a target {target_object} based on a scene description and prior knowledge of the target.

Scenario:
The object detector has identified a scene containing a {target_object}, and the VQA model has provided the following description:

<START_OF_DESCRIPTION>
{distractor_object_description}
<END_OF_DESCRIPTION>

Target object information: 
Based on previous interactions, you know the target picture has the following characteristics:
<START_TARGET_PICTURE_FACTS>
{facts_about_the_target_picture}
<END_TARGET_PICTURE_FACTS> 

Task:
1. Similarity analysis.
Analyze how closely the detected scene description aligns with the known facts about the target {target_object}. Provide a similarity score between 0 and 10, where:
- 0 = The detected {target_object} is not the target object.
- 10 = The detected {target_object} is definitely the target object.
- If no information about the target is available, the score should be -1.

2. Question Generation:
- The question is for the target object, not the detected one.
- Ask exactly one specific, relevant, and human-answerable question related to the target object that maximizes information gain for identifying the target {target_object}.
- Do not ask speculative or irrelevant questions 
- The question should be grounded in observable or known details from the scene, focusing on key characteristics that can help confirm or refute the identity of the target object.

Ensure your response follows the format below:
YAML_START # must be present to get the information back
similarity_score: <similarity score>
questions:
    <question_number>: <question_content>
YAML_END # must be present to get the information back

Provide your reasoning step-by-step for the similarity score and questions, after the YAML_END tag."""
\end{lstlisting}

\section{Algorithm}
\label{supmat:algorithm}
We first present the complete \ourMethod{}'s algorithm in Sec.~\ref{supmat:algorithm_aiuta}.
As outlined in the main paper (Sec.~\ref{sec:method}), \ourMethod{} enriches the zero-shot training policy VLFM~\cite{yokoyama2023vlfm}. 
Specifically, we detail the input/output structure of \ourMethod{} regarding VLFM policy $\pi$, as well as \ourMethod{}'s main component, \ie, the \textit{Self Questioner} (see Sec.~\ref{supmat:algo:self_q}) and the \textit{Interaction Trigger }(Sec.~\ref{supmat:algo_interaction_reasoner}).

\subsection{AIUTA Algorithm}
\label{supmat:algorithm_aiuta}
Algorithm 1 outlines the complete \ourMethod{} pipeline.
Upon detecting a candidate object, \ourMethod{} first invokes the \textit{Self Questioner} module (shown in  Sec.~\ref{supmat:algo:self_q}) to obtain an accurate and detailed understanding of the observed object and to reduce inaccuracies and hallucinations, obtaining a refined observation description $S_{refined}$.
Then, with the known facts about the target instance and the refined description, \ourMethod{} invokes the \textit{Interaction Trigger} module (Sec.~\ref{supmat:algo_interaction_reasoner}) for up to 4 iterations rounds (\ie, Max\_Iteration\_Number $= 4$), as specified in Sec.~\ref{sec:experiments} under \textbf{Implementation Details}.
Within each interaction round, if \ourMethod{} returns the \texttt{STOP} action, then the policy $\pi$ terminates the navigation since the target instance is found; otherwise, the policy $\pi$ continues the navigation process.

\begin{algorithm}[htbp]
\caption{\textbf{AIUTA}}
\begin{algorithmic}[1]
\Require Target object facts $F$, Observation $O_t$, policy $\pi$, Candidate Object Detection, Max Iteration number

\Comment{Upon candidate object detection}
\State $S_{refined} \gets \text{Self\_Questioner}(F, O_t)$
\Comment{enrich details and reduce inaccuracy, obtain a refined description}
\If{ $S_{refined} = \text{``~"}$}
    \State$\pi(\text{CONTINUE\_EXPLORING})$
    \Comment{VQA detection failed, Signal to policy $\pi$ to continue exploration}
\EndIf
\vspace{1mm}
\For{each iteration in Max\_Iteration\_Number}
    \State  $\text{aiuta\_action} \gets \text{Interaction\_Trigger}(F, S_{refined})$
    \If{ $\text{aiuta\_action}$ = STOP}
    \State $\pi(\text{STOP})$
    \Comment{Signal to policy $\pi$ that the object is found! Terminate exploration}
    \Else
    \State$\pi(\text{CONTINUE\_EXPLORING})$
    \Comment{Signal to policy $\pi$ to continue exploration}
    \EndIf
\EndFor

\end{algorithmic}
\end{algorithm}

\newpage
\subsection{Self Questioner}
\label{supmat:algo:self_q}

\begin{algorithm}[htbp]
\caption{\textbf{Self Questioner Module}}

    \begin{algorithmic}[1]
    \Require Target object facts $F$, Uncertainty Threshold $\tau$, Observation $O_t$, $P_{init}, P_{details}, P_{check}, P_{self questions}, P_{refined}$
    
    \vspace{2mm}
    \State \textbf{Step 1: Detailed Detection Description, from $S_{\text{init}}$ to $S_{\text{enriched}}$}
    \State Initial scene description: $S_{\text{init}} \gets \text{VLM}(O_t, P_{\text{init}})$
    \State Self-generate questions to enrich description
    \Statex \hspace{2em} $Q_\atoa^{details} \gets \text{LLM}(P_{\text{details}}, S_{\text{init}}, F)$
    \vspace{1mm}
    
    \For{each question $q_j$ in $Q_\atoa^{details}$}
        \State $r_\atoa \gets \text{VLM}(O_t, q_j)$ \Comment{Get answers}
        \State $S_{\text{init}} \gets \text{concatenate}(S_{\text{init}}, r_\atoa)$
    \EndFor
    \State $S_{\text{enriched}} \gets S_{\text{init}}$ \Comment{Updated scene description}
    
    \vspace{2mm}
    \State \textbf{Step 2: Perception Uncertainty Estimation}
    \State $(r_{\text{check}}, u_{\text{check})} \gets \text{VLM}(O_t, P_{\text{check}})$
    \Comment{Check detection with uncertainty}
    \If{NOT ($r_{\text{check}} = \text{``Yes"}$ AND $u_{\text{check}} = \text{``Certain"}$)}
            \State \textbf{return} ``~"
            \Comment{empty string, thus continue exploring}
    \EndIf
    \State $Q_\atoa^{attribute} \gets \text{LLM}(P_{\text{self questions}}, F, S_{\text{enriched}})$
    \Comment{Generate self-questions to verify attributes}
    \State Container $\gets \{\}$ \Comment{Store question, answer, uncertainty}
    \For{each question $q_j$ in $Q_\atoa^{attribute}$}
        \State $(r_j, u_j) \gets \text{VLM}(O_t, q_j)$ \Comment{Get answers and uncertainties}
        \State Container $\gets \text{concatenate}( \text{Container}, \{q_j, r_j, u_j\})$
    \EndFor
    \vspace{2mm}
    \State \textbf{Step 3: Detection Description Refinement}
    \State $S_{\text{refined}} \gets \text{LLM}(P_{\text{refined}}, \text{Container}, S_{\text{enriched}})$
    \Comment{Filter out uncertain attributes}
    \State \textbf{return} $S_{\text{refined}}$
    \end{algorithmic}
\end{algorithm}

\newpage
\subsection{Interaction Trigger}
\label{supmat:algo_interaction_reasoner}

\begin{algorithm}[htbp]
\caption{\textbf{Interaction Trigger}}
    \begin{algorithmic}[1]
    \Require Target object facts $F$, Refined observation description $S_{\text{refined}}$, $P_{score}$, $\tau_{stop}$ and $\tau_{skip}$
    \State $(s, q_\atou) \gets \text{LLM}(P_{score}, S_{\text{refined}}, F)$
    \Comment{get alignment score $s$, and question for the human $q_\atou$}

    \vspace{2mm}
    \If{$s \geq \tau_{stop}$}
        \State \textbf{return} STOP
        \Comment{target found, stop navigation.}
    \vspace{2mm}
    \ElsIf{$s < \tau_{skip}$}
        \State \textbf{return} CONTINUE\_EXPLORING
        \Comment{skip the question and continue exploring}
    \vspace{2mm}
    \Else
        \State $r_\utoa \gets \text{Ask\_Human}(q_\atou)$
        \Comment{posing clarifying question $q_\atou$ from the agent to the human.}
        \State $F\gets\text{Update\_Facts}(F, r_\utoa)$
        \Comment{update target object facts $F$}
    \EndIf
    \end{algorithmic}
\end{algorithm}

\end{document}